\documentclass{article}
\usepackage{iclr2027_conference,times}

\usepackage{amsmath,amsfonts,bm}

\def\eqref#1{equation~\ref{#1}}

\def\1{\bm{1}}

\DeclareMathAlphabet{\mathsfit}{\encodingdefault}{\sfdefault}{m}{sl}
\SetMathAlphabet{\mathsfit}{bold}{\encodingdefault}{\sfdefault}{bx}{n}

\usepackage{hyperref}
\usepackage{etoc}
\usepackage{url}
\usepackage{booktabs}
\usepackage{array}
\usepackage{graphicx}
\usepackage{wrapfig}
\usepackage{amsmath}
\usepackage{amssymb}
\usepackage{xcolor}
\usepackage{multirow}
\usepackage{placeins}
\usepackage{tikz}
\usetikzlibrary{patterns,positioning,decorations.pathreplacing,calc}

\definecolor{blkP}{HTML}{BBBBBB}
\definecolor{blkA}{HTML}{4477AA}
\definecolor{blkB}{HTML}{EE7733}
\definecolor{blkC}{HTML}{009988}
\definecolor{blkQ}{HTML}{AA3377}
\definecolor{cellon}{HTML}{DDDDDD}
\definecolor{cellx}{HTML}{F4CCCC}
\definecolor{cellfix}{HTML}{99CC99}%

\newcommand{\cls}[1]{\textcircled{\scriptsize #1}}
\newcommand{\cmark}{\checkmark}
\newcommand{\xmark}{$\times$}
\newcommand{\yesmark}{\textcolor{green!50!black}{\checkmark}}
\newcommand{\nomark}{\textcolor{red!65!black}{$\times$}}
\newcommand{\hdr}[1]{\begin{tabular}[b]{@{}c@{}}#1\end{tabular}}
\newcolumntype{P}[1]{>{\raggedright\arraybackslash}p{#1}}
\newcommand{\nacell}{N/A}

\newcommand{\repourlraw}{https://github.com/xishi404/KVShare-Arena}
\newcommand{\leaderboardurlraw}{https://xishi404.github.io/KVShare-Arena/}
\newcommand{\dataseturlraw}{https://github.com/xishi404/KVShare-Arena/tree/main/data/queryset}
\newcommand{\repourl}{\url{\repourlraw}}
\newcommand{\leaderboardurl}{\url{\leaderboardurlraw}}
\newcommand{\dataseturl}{\url{\dataseturlraw}}

\title{KVShareArena: KV-Cache Reuse Across Contexts and Model Checkpoints}

\author{Xi Shi \\
University of Central Florida \\
\texttt{xi320101@ucf.edu}
\And
Mengxin Zheng \\
University of Central Florida \\
\texttt{mengxin.zheng@ucf.edu}
\And
Qian Lou \\
University of Central Florida \\
\texttt{qian.lou@ucf.edu}
}

\iclrfinalcopy

\begin{document}

\maketitle
\lhead{Preprint. Under review at ICLR 2027.}

\begin{abstract}
Reusing key--value (KV) caches speeds up LLM inference by avoiding
repeated computation on shared text. The standard method, prefix caching,
reuses a KV cache only when the LLM is the same and all text before the reused
content is identical. However, real workloads often break both
conditions. The preceding text changes when a
RAG system places different documents before the same document, or when
agents with different system prompts read the same file or tool output. The LLM changes when
multi-agent workflows use specialized LLMs on shared material, or when an
updated model reads documents cached by its previous version. Because KV
caches depend on both the preceding text and the model weights, direct
reuse in these settings can reduce answer quality. Many methods repair the reused cache or compress it, each with its own balance
between quality and cost, so choosing among them requires a comparison
under the same conditions. However, each method paper uses
its own tasks, models, and cost measures, so the reported results cannot
be compared directly. Existing benchmarks do not provide such a comparison either,
because they mainly test long-context processing or reuse of an
unchanged prefix. To fill this gap, we introduce
\textbf{KVShareArena}, a benchmark and open evaluation framework that
compares these methods under the same conditions. KVShareArena has
(1)~reuse tests on 2,150 questions from three QA datasets, in which the
preceding text, the LLM that wrote the cache, or both change while the
answering LLM and the input text stay fixed; (2)~five dense and
mixture-of-experts LLMs from 4B to 30B parameters and six LLM pairs in
which one version of an LLM reads caches written by another, for 33
model--dataset settings in total; (3)~11 repair and compression methods
from six method classes; (4)~four evaluation perspectives: answer
quality, prefill computation, KV-cache memory, and latency; and (5)~a
common interface for adding new methods and an interactive leaderboard
that lets users set their own balance between quality and speed.
Experiments on KVShareArena yield two findings. First, both the quality loss from reuse and which repairs help depend on the LLM, even between two 8B models. Second, most repairs keep their quality when
another version of the LLM wrote the cache, but a trained repair adapter
loses quality in 12 of the 18 pair--dataset tests.
Code and data are available at \repourl.
\end{abstract}

\etocdepthtag.toc{mtmain}
\section{Introduction}
\label{sec:intro}

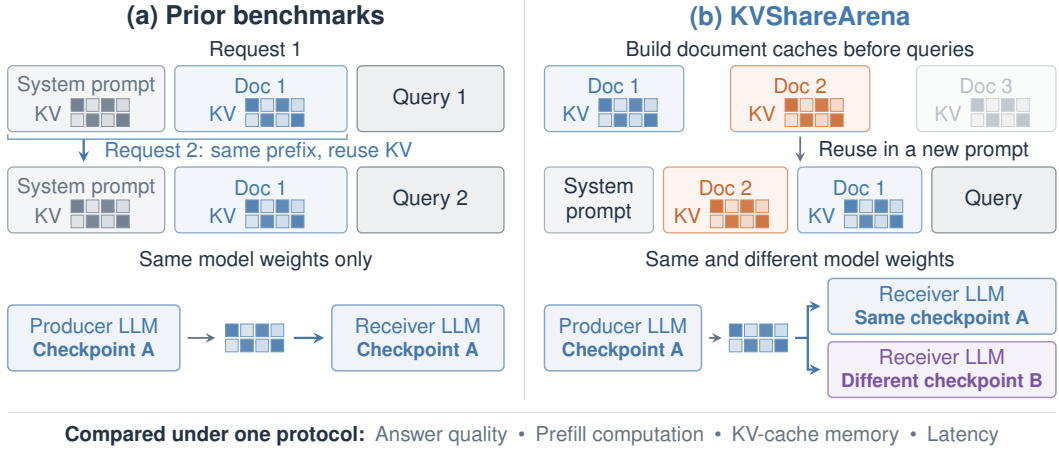
\begin{figure}[t]
\centering
\resizebox{\linewidth}{!}{\begin{tikzpicture}[x=1pt,y=-1pt]
\definecolor{wallA}{HTML}{4477AA}
\definecolor{wallB}{HTML}{EE7733}
\definecolor{wallInk}{HTML}{243344}
\definecolor{wallMuted}{HTML}{667382}
\definecolor{wallPurple}{HTML}{7954A2}
\colorlet{wallDocTwo}{wallB!85!black}
\colorlet{wallUnused}{wallMuted!45}
\tikzset{
  wall label/.style={font=\sffamily\fontsize{8.3}{9.6}\selectfont,
                    text=wallInk,inner sep=0pt},
  wall title/.style={wall label,font=\sffamily\bfseries\fontsize{10.2}{11.5}\selectfont,
                    align=center},
  wall small/.style={wall label,font=\sffamily\fontsize{8.1}{9.3}\selectfont},
  wall note/.style={wall small,font=\sffamily\itshape\fontsize{8.1}{9.3}\selectfont,
                   text=wallMuted,align=center},
  wall arrow/.style={-stealth,line width=.7pt,draw=wallMuted},
  wall reuse/.style={-stealth,line width=.9pt,draw=wallA}}
\newcommand{\wallCells}[3]{%
  \foreach \shade [count=\cell from 0] in {90,20,85,25} {
    \filldraw[fill=#3!\shade!white,draw=#3!75,line width=.25pt]
      ({#1+6*\cell},{#2}) rectangle ({#1+6*\cell+5},{#2+5});}
  \foreach \shade [count=\cell from 0] in {20,85,25,90} {
    \filldraw[fill=#3!\shade!white,draw=#3!75,line width=.25pt]
      ({#1+6*\cell},{#2+6}) rectangle ({#1+6*\cell+5},{#2+11});}}
\newcommand{\wallCacheCard}[5]{%
  \filldraw[rounded corners=2pt,fill=#4!7,draw=#4!65,line width=.5pt]
    ({#1},{#2}) rectangle ({#1+#3},{#2+26});
  \node[wall label,text=#4,align=center,font=\sffamily\fontsize{8.1}{9.3}\selectfont]
    at ({#1+#3/2},{#2+8}) {#5};
  \node[wall small,text=#4] at ({#1+#3/2-15},{#2+19}) {KV};
  \wallCells{#1+#3/2-6}{#2+13}{#4}}
\newcommand{\wallBox}[7]{%
  \filldraw[rounded corners=2pt,fill=#5!8,draw=#5!55,line width=.5pt]
    ({#1},{#2}) rectangle ({#1+#3},{#2+#4});
  \node[wall label,text=#6,align=center]
    at ({#1+#3/2},{#2+#4/2}) {#7};}
\newcommand{\wallModel}[7]{%
  \filldraw[rounded corners=2pt,fill=#5!8,draw=#5!65,line width=.6pt]
    ({#1},{#2}) rectangle ({#1+#3},{#2+#4});
  \node[wall small,text=#5] at ({#1+#3/2},{#2+#4/2-5}) {#6};
  \node[wall label,text=#5,font=\sffamily\bfseries\fontsize{7.5}{8.6}\selectfont]
    at ({#1+#3/2},{#2+#4/2+5}) {#7};}

\node[wall title] at (98,7) {(a) Prior benchmarks};
\node[wall small] at (98,20) {Request 1};
\wallCacheCard{0}{26}{62}{wallMuted}{System prompt}
\wallCacheCard{66}{26}{68}{wallA}{Doc 1}
\wallBox{138}{26}{58}{26}{wallInk}{wallInk}{Query 1}
\draw[wallA!70,line width=.6pt] (0,53) -- (0,55) -- (134,55) -- (134,53);
\draw[wall reuse] (30,55) -- (30,64);
\node[wall small,text=wallA,anchor=west] at (38,60) {Request 2: same prefix, reuse KV};
\wallCacheCard{0}{66}{62}{wallMuted}{System prompt}
\wallCacheCard{66}{66}{68}{wallA}{Doc 1}
\wallBox{138}{66}{58}{26}{wallInk}{wallInk}{Query 2}

\draw[wallInk!15,line width=.5pt] (204,0) -- (204,158);

\node[wall title,text=wallA] at (313,7) {(b) KVShareArena};
\node[wall small] at (313,20) {Build document caches before queries};
\wallCacheCard{212}{26}{55}{wallA}{Doc 1}
\wallCacheCard{285.5}{26}{55}{wallDocTwo}{Doc 2}
\wallCacheCard{359}{26}{55}{wallUnused}{Doc 3}
\draw[wall arrow] (313,54) -- (313,64);
\node[wall small,anchor=west] at (320,59.5) {Reuse in a new prompt};
\wallBox{212}{66}{43}{26}{wallMuted}{wallInk}{System\\prompt}
\wallCacheCard{259}{66}{49}{wallDocTwo}{Doc 2}
\wallCacheCard{312}{66}{49}{wallA}{Doc 1}
\wallBox{365}{66}{49}{26}{wallInk}{wallInk}{Query}

\node[wall small] at (98,103) {Same model weights only};
\wallModel{0}{120.5}{68}{26}{wallA}{Producer LLM}{Checkpoint A}
\draw[wall arrow] (71,133.5) -- (82,133.5);
\wallCells{86}{128}{wallA}
\draw[wall reuse] (113,133.5) -- (125,133.5);
\wallModel{128}{120.5}{68}{26}{wallA}{Receiver LLM}{Checkpoint A}

\node[wall small] at (313,103) {Same and different model weights};
\wallModel{212}{120.5}{62}{26}{wallA}{Producer LLM}{Checkpoint A}
\draw[wall arrow] (277,133.5) -- (282,133.5);
\wallCells{285}{128}{wallA}
\draw[wallA,line width=.9pt] (311,133.5) -- (316,133.5);
\draw[wall reuse] (316,133.5) -- (316,121) -- (321,121);
\draw[wall reuse] (316,133.5) -- (316,146) -- (321,146);
\wallModel{324}{110}{90}{22}{wallA}{Receiver LLM}{Same checkpoint A}
\wallModel{324}{135}{90}{22}{wallPurple}{Receiver LLM}{Different checkpoint B}

\draw[wallInk!15,line width=.5pt] (0,163) -- (414,163);
\node[wall small,align=center,text=wallMuted] at (207,172)
  {\textbf{\textcolor{wallInk}{Compared under one protocol:}}\enspace Answer quality\enspace\textbullet\enspace
   Prefill computation\enspace\textbullet\enspace KV-cache memory\enspace\textbullet\enspace Latency};
\path[use as bounding box] (0,0) rectangle (414,177);
\end{tikzpicture}}
\caption{\textbf{KVShareArena evaluates reuse after context or model
weights change.} \textbf{(a)}~SCBench \citep{scbench2025} tests reuse of an
unchanged prefix with the same model weights. \textbf{(b)}~KVShareArena
combines separately built document caches in new prompts and reads them
with the same checkpoint~A or a different checkpoint~B of the same
architecture and tokenizer. Small grids
represent KV caches.}
\label{fig:wall}
\end{figure}

\begin{wrapfigure}{R}{0.43\textwidth}
\vspace{-\intextsep}
\centering
\resizebox{\linewidth}{!}{\begin{tikzpicture}[x=1pt,y=-1pt]
\definecolor{shareBlue}{HTML}{4477AA}
\definecolor{shareOrange}{HTML}{EE7733}
\definecolor{shareInk}{HTML}{243344}
\definecolor{shareMuted}{HTML}{667382}
\definecolor{shareRepair}{HTML}{7954A2}
\definecolor{shareWarning}{HTML}{AA3344}
\tikzset{
  share label/.style={font=\sffamily\fontsize{9.2}{10.4}\selectfont,
                     text=shareInk,inner sep=0pt},
  share title/.style={share label,font=\sffamily\bfseries\fontsize{10.2}{11.4}\selectfont,
                     anchor=west},
  share note/.style={share label,font=\sffamily\fontsize{9.2}{10.4}\selectfont,
                    align=center},
  share action/.style={share label,font=\sffamily\fontsize{8.6}{9.7}\selectfont,
                      text=shareMuted},
  share arrow/.style={-stealth,line width=.85pt,draw=shareMuted}}

\newcommand{\shareText}[5]{%
  \filldraw[fill=#4!10,draw=#4!65,line width=.45pt]
    ({#1},{#2}) rectangle ({#1+#3},{#2+14});
  \node[share label,text=#4!75!black] at ({#1+#3/2},{#2+7}) {#5};}
\newcommand{\shareKV}[6]{%
  \filldraw[fill=#4!24,draw=#4!65,line width=.4pt]
    ({#1},{#2}) -- ({#1+5},{#2-5}) -- ({#1+#3+5},{#2-5})
    -- ({#1+#3},{#2}) -- cycle;
  \filldraw[fill=#4!42,draw=#4!65,line width=.4pt]
    ({#1+#3},{#2}) -- ({#1+#3+5},{#2-5})
    -- ({#1+#3+5},{#2+9}) -- ({#1+#3},{#2+14}) -- cycle;
  \filldraw[fill=#4!36,draw=#4!65,line width=.4pt]
    ({#1},{#2}) rectangle ({#1+#3},{#2+14});
  \ifnum#6=1\relax
    \begin{scope}
      \clip ({#1},{#2}) rectangle ({#1+#3},{#2+14});
      \foreach \stripe in {-16,-10,...,50} {
        \draw[shareInk!50,line width=.4pt]
          ({#1+\stripe},{#2+14}) -- ({#1+\stripe+14},{#2});}
    \end{scope}
  \fi
  \ifnum#6=2\relax
    \foreach \stripe in {5,15} {
      \fill[shareRepair!85] ({#1+\stripe},{#2})
        rectangle ({#1+\stripe+4},{#2+14});
      \fill[shareRepair!50] ({#1+\stripe},{#2})
        -- ({#1+\stripe+5},{#2-5}) -- ({#1+\stripe+9},{#2-5})
        -- ({#1+\stripe+4},{#2}) -- cycle;}
  \fi
  \if\relax\detokenize{#5}\relax\else
  \node[share label,font=\sffamily\bfseries\fontsize{8.8}{10}\selectfont,
        fill=white,fill opacity=.82,text opacity=1,inner sep=.6pt]
    at ({#1+#3/2},{#2+7}) {#5};\fi}
\newcommand{\shareFlow}[2]{%
  \node[share action] at (119,{#1+19}) {#2};
  \draw[share arrow] (106,{#1+28}) -- (130,{#1+28});}
\newcommand{\shareQuestion}[1]{%
  \shareText{64}{#1+21}{38}{shareMuted}{Question}}
\newcommand{\shareOutput}[3]{%
  \shareKV{136}{#1+21}{26}{shareBlue}{}{#2}
  \shareKV{162}{#1+21}{26}{shareOrange}{}{#3}
  \shareKV{188}{#1+21}{26}{shareMuted}{}{1}}

\node[share title] at (0,6) {(a) Full recomputation};
\shareText{0}{21}{26}{shareBlue}{Doc 1}
\shareText{32}{21}{26}{shareOrange}{Doc 2}
\shareQuestion{0}
\shareFlow{0}{Slowest}
\shareOutput{0}{1}{1}
\node[share note] at (111,43) {Full prefill\enspace\textbullet\enspace Quality reference};

\node[share title] at (0,55) {(b) Prefix reuse};
\shareKV{0}{70}{26}{shareBlue}{KV 1}{0}
\shareText{32}{70}{26}{shareOrange}{Doc 2}
\shareQuestion{49}
\shareFlow{49}{Faster}
\shareOutput{49}{0}{1}
\node[share note] at (111,92) {Less prefill\enspace\textbullet\enspace Same quality};

\node[share title] at (0,104) {(c) Direct cross-context reuse};
\shareKV{0}{119}{26}{shareBlue}{KV 1}{0}
\shareKV{32}{119}{26}{shareOrange}{KV 2}{0}
\shareQuestion{98}
\shareFlow{98}{Fastest}
\shareOutput{98}{0}{0}
\node[share label,font=\sffamily\bfseries\fontsize{8}{9.2}\selectfont,text=shareWarning,
      fill=white,fill opacity=.85,text opacity=1,inner sep=1pt] at (175,126) {No cross-attention};
\node[share note,text=shareWarning] at (111,142) {Quality can drop};

\node[share title] at (0,153) {(d) Reuse with repair};
\shareKV{0}{168}{26}{shareBlue}{KV 1}{0}
\shareKV{32}{168}{26}{shareOrange}{KV 2}{0}
\shareQuestion{147}
\shareFlow{147}{Faster?}
\shareOutput{147}{0}{2}
\node[share note,font=\sffamily\bfseries\fontsize{9.2}{10.4}\selectfont]
  at (111,190) {Quality recovered? At what cost?};

\filldraw[fill=shareMuted!12,draw=shareMuted!65,line width=.4pt]
  (1,199) rectangle (9,206);
\begin{scope}
  \clip (1,199) rectangle (9,206);
  \foreach \stripe in {-6,-2,...,10} {
    \draw[shareInk!50,line width=.4pt]
      (\stripe,206) -- ({\stripe+7},199);}
\end{scope}
\node[share note,anchor=west,text=shareMuted,
      font=\sffamily\fontsize{8.8}{10}\selectfont]
  at (14,202.5) {Computed in this request};
\filldraw[fill=shareOrange!36,draw=shareOrange!65,line width=.4pt]
  (126,199) rectangle (134,206);
\fill[shareRepair!85] (129,199) rectangle (132,206);
\node[share note,anchor=west,text=shareMuted,
      font=\sffamily\fontsize{8.8}{10}\selectfont]
  at (139,202.5) {Modified by repair};
\path[use as bounding box] (0,0) rectangle (224,210);
\end{tikzpicture}}
\caption{\textbf{Reuse saves prefill but can lose quality.} Repair tries
to recover it at extra cost.}
\label{fig:kv-sharing}
\end{wrapfigure}
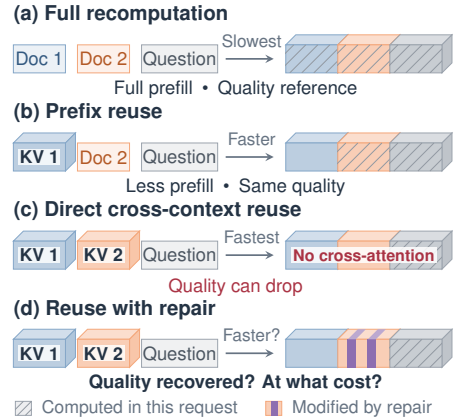
LLM applications repeatedly process shared text, including instructions,
documents, and conversation histories. In retrieval-augmented generation
(RAG), for example, many queries retrieve the same documents. Before
answering, the model encodes the text into key--value (KV) caches during
\emph{prefill}. Repeating prefill on the same text adds computation and
delays the first output token. Reusing KV caches computed earlier avoids
this repeated work \citep{epic2025,cacheblend2025}.

Figure~\ref{fig:kv-sharing}, adapted from \citet{cacheblend2025},
shows four ways to process a RAG prompt.
The standard method, prefix caching, avoids recomputing an unchanged prefix (Figure~\ref{fig:kv-sharing}b). It is used in vLLM and SGLang
\citep{radixattention2024} and offered by commercial LLM APIs.
Cached prefixes include system prompts, tool descriptions, and documents.
However, reuse needs an exact match from the start of the prompt through the cached text (Figure~\ref{fig:wall}a).

\paragraph{Cross-context reuse.}
Many LLM applications build each prompt from reusable pieces of text,
such as documents, few-shot examples, and tool descriptions
\citep{promptcache2024,epic2025,minipic2026}. We call each such piece a
\emph{block}. The same block often recurs with different text before it
\citep{epic2025}. In RAG, the same document is retrieved together with different documents for different queries \citep{cacheblend2025,cachecraft2025}; in multi-agent systems,
the same file or tool output follows a different system prompt in each
agent \citep{kvcomm2025,minipic2026}. Because it needs an exact match, prefix caching cannot reuse such a block once its preceding text changes. In practice, exact prefix caching could serve only 8\% of requests in one production RAG service, and 75\% of the
blocks retrieved for each query had to be processed again
\citep{cachecraft2025}. To reuse such blocks, recent methods encode each block separately and reuse its cache after different preceding text
\citep{epic2025,hypic2026}; we call this \emph{cross-context reuse}
(Figure~\ref{fig:wall}b). A separately encoded block, however, has not
seen its new preceding text. In Figure~\ref{fig:kv-sharing}c, Doc~2's KV was computed without Doc~1, so direct reuse, which uses the cache as is, can lower answer quality.

\paragraph{Cross-checkpoint reuse.}
Shared text also passes between different LLMs. For example, multi-agent workflows use models fine-tuned for different tasks. In DroidSpeak, a coding agent and a testing agent run different fine-tuned LLMs, and the testing agent must prefill the task instructions and code that the coding agent has already processed \citep{droidspeak2025}. Likewise, an updated model prefills again the documents its previous version has already cached. Prefix caching does not help here: it cannot reuse a cache written by another LLM, even for the same text.
We call the LLM that writes a cache the \emph{producer} and the LLM that
reads it the \emph{receiver}. We call reuse between a producer and a
receiver with different weights (different checkpoints) \emph{cross-checkpoint reuse}
(Figure~\ref{fig:wall}b). Different weights, however, produce different KV values for the same text, even with the same architecture and tokenizer. Direct reuse can therefore lower answer quality
\citep{droidspeak2025}.

In both settings, the reused text is unchanged. Yet \textbf{the same
text does not guarantee reusable KV.} When the preceding text changes, a reused cache misses that text and also sits at new positions in the prompt. The simplest fix, \emph{position alignment}, re-encodes the cache's positions to match its new place in the prompt, without recomputing any token. It fixes the positions, but it does not add the missing context
or adapt the cache to new weights. Such fixes, which we call \emph{repair}, take extra computation when the cache is reused, or training the model or an adapter in advance. Because repair adds cost, reuse is useful only if it preserves enough answer quality and still costs less than recomputation
(Figure~\ref{fig:kv-sharing}d).

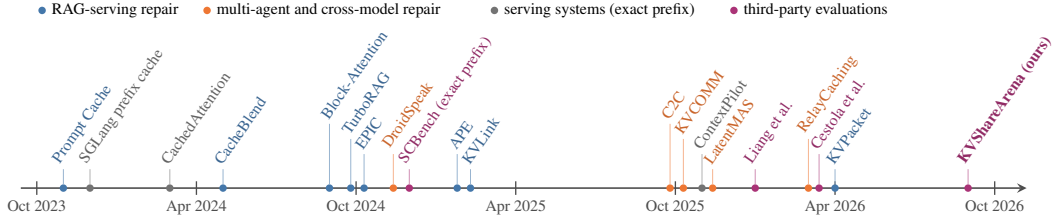
\begin{figure}[t]
\centering
\begin{tikzpicture}[x=0.352cm,y=1cm]
\colorlet{tlrag}{blkA}
\colorlet{tlmas}{blkB}
\colorlet{tlserv}{black!55}
\colorlet{tleval}{blkQ}
\newcommand{\up}[4]{%
  \fill[#2] (#1,0) circle (1.4pt);
  \draw[#2!55,line width=0.35pt] (#1,0.05) -- (#1,#4);
  \node[font=\tiny,rotate=55,anchor=south west,inner sep=0.8pt,text=#2!85!black] at (#1,#4) {#3};}
\draw[-stealth,line width=0.6pt,black!70] (-0.6,0) -- (37.2,0);
\foreach \m/\lab in {0/Oct 2023, 6/Apr 2024, 12/Oct 2024, 18/Apr 2025, 24/Oct 2025, 30/Apr 2026, 36/Oct 2026}
  {\draw[black!70,line width=0.5pt] (\m,0.09) -- (\m,-0.09);
   \node[font=\tiny,text=black!70,anchor=north,inner sep=1.5pt] at (\m,-0.1) {\lab};}
\up{1}{tlrag}{Prompt Cache}{0.25}
\up{2}{tlserv}{SGLang prefix cache}{0.25}
\up{5}{tlserv}{CachedAttention}{0.25}
\up{7}{tlrag}{CacheBlend}{0.25}
\up{11}{tlrag}{Block-Attention}{0.85}
\up{11.8}{tlrag}{TurboRAG}{0.65}
\up{12.3}{tlrag}{EPIC}{0.45}
\up{13.4}{tlmas}{DroidSpeak}{0.45}
\up{14}{tleval}{SCBench (exact prefix)}{0.25}
\up{15.8}{tlrag}{APE}{0.45}
\up{16.3}{tlrag}{KVLink}{0.25}
\up{23.8}{tlmas}{C2C}{0.85}
\up{24.3}{tlmas}{KVCOMM}{0.65}
\up{25}{tlserv}{ContextPilot}{0.45}
\up{25.4}{tlmas}{LatentMAS}{0.25}
\up{27}{tleval}{Liang et al.}{0.25}
\up{29}{tlmas}{RelayCaching}{0.65}
\up{29.4}{tleval}{Cestola et al.}{0.45}
\up{30}{tlrag}{KVPacket}{0.25}
\up{35}{tleval}{\textbf{KVShareArena (ours)}}{0.25}
\begin{scope}[shift={(0.2,2.35)}]
\foreach \dx/\col/\txt in {0/tlrag/RAG-serving repair, 6.2/tlmas/multi-agent and cross-model repair, 17/tlserv/serving systems (exact prefix), 26/tleval/third-party evaluations}
  {\fill[\col] (\dx,0) circle (1.4pt); \node[font=\tiny,anchor=west,inner sep=1.5pt] at (\dx+0.2,0) {\txt};}
\end{scope}
\end{tikzpicture}
\caption{Repair methods for reused KV caches have appeared at a steady pace since late 2023, in RAG serving and in multi-agent systems, while third-party evaluation lags behind: one benchmark that reuses caches only under an exact prefix and two studies of cross-context reuse in 2026. Dates are arXiv first versions.}
\label{fig:timeline}
\end{figure}

\paragraph{A missing benchmark setting.}
Several lines of work repair reused caches (Figure~\ref{fig:timeline}).
RAG serving systems recompute a small share of tokens
\citep{cacheblend2025,epic2025} or recalibrate attention across cached
documents \citep{ape2025}. Other methods train the model or a small
adapter to read separately cached blocks
\citep{blockattention2024,kvpacket2026}, and multi-agent systems correct
caches with offsets stored from earlier requests \citep{kvcomm2025}.
However, each study uses its own tasks, models, quality baselines, and cost measures, so a gain reported in one study may reflect a better method, an
easier setting, or a different way of measuring. Existing benchmarks and studies do not compare these methods under the same conditions (Table~\ref{tab:landscape}). SCBench tests only reuse of an
unchanged prefix with the same weights \citep{scbench2025}. Two studies compare cross-context methods, but only on quality and not across checkpoints
\citep{elevenmethodstudy2026,kvreusefails2026}. DroidSpeak builds a
system for cross-checkpoint reuse rather than comparing methods \citep{droidspeak2025}. The open
question is: \textbf{Which methods preserve answer quality at a lower
cost than recomputation after context or checkpoint changes?}

\begin{table}[t]
\caption{\textbf{Of the evaluations listed, only KVShareArena covers
reuse across both contexts and checkpoints, all four evaluation
perspectives, and an open leaderboard.} SCBench compares compute and
memory complexity; Liang et al.\ also report selection consistency and
reuse rate.}
\label{tab:landscape}
\centering
\small
\setlength{\tabcolsep}{4.5pt}
\begin{tabular}{@{}lccccccc@{}}
\toprule
& \multicolumn{2}{c}{Reuse across} & \multicolumn{4}{c}{Evaluation perspectives} & Open \\
\cmidrule(lr){2-3}\cmidrule(lr){4-7}
Evaluation & Contexts & Checkpoints & Quality & Compute & Memory & Latency & leaderboard \\
\midrule
SCBench & \nomark & \nomark & \yesmark & \yesmark & \yesmark & \nomark & \nomark \\
Cestola et al.\ (2026) & \yesmark & \nomark & \yesmark & \nomark & \nomark & \nomark & \nomark \\
Liang et al.\ (2026) & \yesmark & \nomark & \yesmark & \nomark & \nomark & \nomark & \nomark \\
\midrule
\textbf{KVShareArena} (ours) & \yesmark & \yesmark & \yesmark & \yesmark & \yesmark & \yesmark & \yesmark \\
\bottomrule
\end{tabular}
\end{table}

Answering this question needs three design choices. First, the preceding text and the producer weights both change the KV, so they must be varied separately. Second, quality must be measured against full recomputation over the same text, so that a loss reflects reuse, not missing information. Third, each cost must be measured on its own, because one saving does not imply another: a method can avoid prefill computation yet keep the full cache.

We introduce \textbf{KVShareArena}, a benchmark and open evaluation
framework for cross-context and cross-checkpoint reuse of document KV
caches in RAG, built on these three choices. We measure quality by performance-gap recovered (PGR): $0$ matches answering without the documents, and $1$ matches full recomputation over the same text. Experiments on five LLMs from 4B to 30B yield two findings. First, the
quality loss from reuse and the repairs that help depend on the model:
position alignment alone reaches a multi-hop PGR of $.41$ on Llama-3.1-8B
but $.03$ on Qwen3-8B, and CacheBlend, which recomputes a small share of tokens, significantly improves multi-hop QA
on Qwen3-8B but not on Llama-3.1-8B (\S\ref{sec:confirmation}). Second,
most repairs change by less than $.02$ F1 on average when another version
of the LLM wrote the cache, while KVPacket's trained adapter loses up to
$.146$ F1. Retraining the adapter on the new producer's caches recovers this loss, at about $20$ GPU-hours per producer with Qwen3-8B as the receiver (\S\ref{sec:crosscheckpoint}). The experiments also show that compression, which drops or quantizes cached tokens, saves $25$--$72\%$ of KV memory, but its best multi-hop PGR is $.27$, against $.44$--$.52$ for
the leading repairs (\S\ref{sec:re-board}).

Our contributions are:
\begin{itemize}
\setlength{\itemsep}{1pt}
\item \textbf{A controlled protocol.} KVShareArena holds the receiver and
the input text fixed and varies the preceding text and the
producer weights separately, so each source of quality loss is measured
on its own (\S\ref{sec:factorial}). Every method uses the same fixed questions, texts, prompts, and scoring rules.
\item \textbf{Four evaluation perspectives.} KVShareArena measures quality
as PGR, and prefill computation, KV-cache memory, and latency as
recomputed tokens, cache bytes, and time to first token
(\S\ref{sec:frontiers}). The leaderboard ranks all methods by quality and
speed (TTFT saved) and reports memory separately.
\item \textbf{An open framework.} A new method implements three calls and
is evaluated under the same protocol. Researchers can submit its
per-sample outputs for review and inclusion in the public leaderboard
(Figure~\ref{fig:overview}; Appendix~\ref{sec:framework}).
\end{itemize}

\section{KVShareArena}
\label{sec:setup}

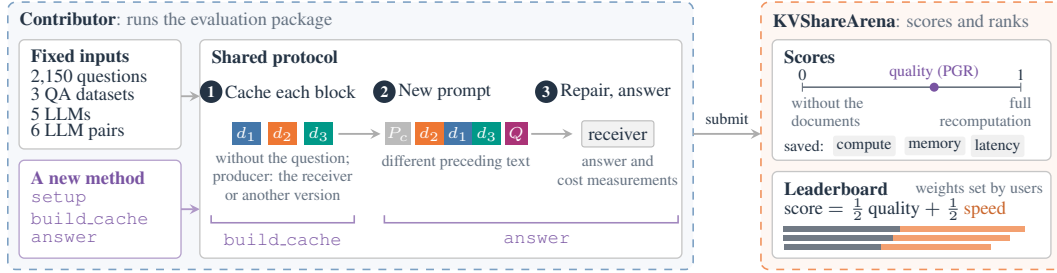
\begin{figure}[t]
\centering
\definecolor{ovInk}{HTML}{243344}
\definecolor{ovMuted}{HTML}{667382}
\definecolor{ovRepair}{HTML}{7954A2}
\resizebox{\linewidth}{!}{%
\begin{tikzpicture}[
  x=1cm, y=1cm, font=\scriptsize,
  zone/.style={draw, dashed, rounded corners=3pt, line width=.6pt},
  card/.style={draw=gray!45, fill=white, rounded corners=2pt, line width=.5pt},
  ttl/.style={font=\scriptsize\bfseries, text=ovInk, inner sep=0pt},
  txt/.style={font=\scriptsize, text=ovInk, inner sep=0pt, align=left},
  sub/.style={font=\fontsize{6}{7}\selectfont, text=ovMuted, inner sep=0pt, align=center},
  code/.style={font=\scriptsize\ttfamily, text=ovRepair, inner sep=0pt, anchor=west},
  chip/.style={font=\fontsize{6}{7}\selectfont, text=ovInk, fill=gray!12, rounded corners=1pt, inner sep=1.5pt, anchor=west},
  num/.style={circle, fill=ovInk, text=white, font=\fontsize{6}{6}\selectfont\bfseries, inner sep=0pt, minimum size=8.5pt},
  blk/.style={minimum width=.38cm, minimum height=.30cm, inner sep=0pt, text=white, font=\fontsize{6}{6}\selectfont},
  arr/.style={->, >=stealth, gray!65, line width=.7pt}]

\draw[zone, draw=blkA!70, fill=blkA!5] (0,0) rectangle (9.1,3.55);
\draw[zone, draw=blkB!80, fill=blkB!6] (10.0,0) rectangle (14,3.55);
\node[ttl, anchor=west] at (.15,3.3) {Contributor\textcolor{ovMuted}{\normalfont: runs the evaluation package}};
\node[ttl, anchor=west] at (10.15,3.3) {KVShareArena\textcolor{ovMuted}{\normalfont: scores and ranks}};

\draw[card] (.15,1.6) rectangle (2.3,3.05);
\node[ttl, anchor=west] at (.3,2.82) {Fixed inputs};
\node[txt, anchor=west] at (.3,2.55) {2,150 questions};
\node[txt, anchor=west] at (.3,2.31) {3 QA datasets};
\node[txt, anchor=west] at (.3,2.07) {5 LLMs};
\node[txt, anchor=west] at (.3,1.83) {6 LLM pairs};

\draw[card, draw=ovRepair!55] (.15,.15) rectangle (2.3,1.45);
\node[ttl, anchor=west, text=ovRepair] at (.3,1.22) {A new method};
\node[code] at (.3,.94) {setup};
\node[code] at (.3,.68) {build\_cache};
\node[code] at (.3,.42) {answer};

\draw[arr] (2.3,2.3) -- (2.55,2.3);
\draw[arr, ovRepair!60] (2.3,.8) -- (2.55,.8);

\draw[card] (2.55,.15) rectangle (8.95,3.05);
\node[ttl, anchor=west] at (2.7,2.8) {Shared protocol};
\def\ca{3.65} \def\cb{5.95} \def\cc{8.1}
\node[num] at (\ca-.95,2.36) {1};  \node[txt, anchor=west] at (\ca-.77,2.36) {Cache each block};
\node[num] at (\cb-.95,2.36) {2};  \node[txt, anchor=west] at (\cb-.77,2.36) {New prompt};
\node[num] at (\cc-.95,2.36) {3};  \node[txt, anchor=west] at (\cc-.77,2.36) {Repair, answer};

\node[blk, fill=blkA] at (\ca-.48,1.8) {$d_1$};
\node[blk, fill=blkB] at (\ca,1.8) {$d_2$};
\node[blk, fill=blkC] at (\ca+.48,1.8) {$d_3$};
\node[sub] at (\ca,1.24) {without the question;\\producer: the receiver\\or another version};
\node[blk, fill=blkP, minimum width=.34cm] at (\cb-.77,1.8) {$P_c$};
\node[blk, fill=blkB] at (\cb-.35,1.8) {$d_2$};
\node[blk, fill=blkA] at (\cb+.03,1.8) {$d_1$};
\node[blk, fill=blkC] at (\cb+.41,1.8) {$d_3$};
\node[blk, fill=blkQ, minimum width=.30cm] at (\cb+.8,1.8) {$Q$};
\node[sub] at (\cb,1.42) {different preceding text};
\node[draw=gray!55, fill=gray!10, rounded corners=1pt, inner sep=2.5pt, font=\scriptsize, text=ovInk] at (\cc,1.8) {receiver};
\node[sub] at (\cc,1.32) {answer and\\cost measurements};
\draw[arr] (\ca+.74,1.8) -- (\cb-1.0,1.8);
\draw[arr] (\cb+1.0,1.8) -- (\cc-.6,1.8);

\draw[ovRepair!70, line width=.6pt] (\ca-.95,.72) -- ++(0,-.1) -- (\ca+.95,.62) -- ++(0,.1);
\node[code, anchor=center] at (\ca,.4) {build\_cache};
\draw[ovRepair!70, line width=.6pt] (\cb-.95,.72) -- ++(0,-.1) -- (\cc+.8,.62) -- ++(0,.1);
\node[code, anchor=center] at ({(\cb+\cc)/2},.4) {answer};

\draw[arr] (9.1,1.8) -- (10.0,1.8);
\node[sub, text=ovInk] at (9.55,2.0) {submit};

\draw[card] (10.15,1.45) rectangle (13.85,3.05);
\node[ttl, anchor=west] at (10.3,2.82) {Scores};
\draw[ovInk!70, line width=.6pt] (10.55,2.42) -- (13.45,2.42);
\foreach \x/\l in {10.55/0, 13.45/1} {\draw[ovInk!70] (\x,2.36) -- (\x,2.48); \node[sub, text=ovInk, anchor=south] at (\x,2.51) {\l};}
\fill[ovRepair] (12.3,2.42) circle (1.7pt);
\node[sub, text=ovRepair, anchor=south] at (12.3,2.51) {quality (PGR)};
\node[sub, anchor=north west, align=left] at (10.4,2.3) {without the\\documents};
\node[sub, anchor=north east, align=right] at (13.6,2.3) {full\\recomputation};
\node[sub, text=ovInk, anchor=west] at (10.3,1.64) {saved:};
\node[chip] at (10.95,1.64) {compute};
\node[chip] at (11.9,1.64) {memory};
\node[chip] at (12.78,1.64) {latency};

\draw[card] (10.15,.15) rectangle (13.85,1.3);
\node[ttl, anchor=west] at (10.3,1.07) {Leaderboard};
\node[sub, anchor=east] at (13.75,1.07) {weights set by users};
\node[txt, anchor=west] at (10.3,.8) {score $=\tfrac12$\,\textcolor{ovInk}{quality} $+\ \tfrac12$\,\textcolor{blkB!85!black}{speed}};
\foreach \i/\q/\v in {0/1.55/1.65, 1/1.45/1.55, 2/1.3/1.45} {
  \fill[ovInk!65] (10.3,.5-\i*.12) rectangle ++(\q,.075);
  \fill[blkB!70] (10.3+\q,.5-\i*.12) rectangle ++(\v,.075);}
\end{tikzpicture}%
}
\caption{\textbf{KVShareArena needs three calls from a new method and
supplies the inputs, protocol, and scoring.} Purple marks the method's own
code.}
\label{fig:overview}
\end{figure}

Figure~\ref{fig:overview} shows how KVShareArena evaluates a method. This
section follows the figure: the data and models (\S\ref{sec:data}), how
caches are built and reused (\S\ref{sec:factorial}), and how answers and
costs are scored (\S\ref{sec:pgr}).

\subsection{Data and models}
\label{sec:data}
\label{sec:tracks}
The main evaluation caches document blocks separately and assembles
them for each question. It uses three
QA datasets in the LongBench~v1 format \citep{longbench2024}: Qasper,
MultiFieldQA, and HotpotQA, which span scientific-paper, single-document,
and multi-hop QA. HotpotQA blocks are separate Wikipedia articles, as in
multi-document retrieval. Qasper and MultiFieldQA blocks are 512-token
windows of one long document, as when a RAG system splits a document into
chunks. Every method reads the same documents the dataset provides for
each question, so retrieval quality does not affect the comparison. We
extend LongBench's Qasper and HotpotQA subsets with more questions from
the original datasets, to $1{,}000$ questions each; MultiFieldQA keeps
its $150$ (Appendix~\ref{app:data}). The documents of each question form
$2$ to $43$ separately cached blocks. Combining several separately cached
blocks is how RAG builds a prompt, and it is where reuse loses quality
and repairs differ (Appendix~\ref{sec:findings}).
Appendix~\ref{app:agent-reports} runs the same test on reports written
by specialist agents (\emph{Agent Reports}).

We evaluate five models. Each model is both the
producer and the receiver of its own caches: Qwen3-8B, Qwen3-4B,
Qwen3-14B, the mixture-of-experts Qwen3-30B-A3B-Instruct-2507, and
Llama-3.1-8B-Instruct. Six pairs, each joining a receiver with another
version of the same model, test a producer change
(\S\ref{sec:crosscheckpoint}).

\subsection{Context and producer changes}
\label{sec:factorial}
\label{sec:protocol}
We vary two factors independently: whether a block is read after the same
text it was encoded with, and whether the producer weights match the
receiver weights. The resulting $2\!\times\!2$ design has four cells: the context is the same or different, and the weights are the same or different. In the
same-context cells the producer prefills the whole prompt in order and
the receiver computes only the question; with matching weights this is
exact prefix caching, which loses nothing and serves as a control.

In the different-context cells, each block is encoded alone, from position zero, with no text before it. The caches of $m$ blocks are then assembled into one receiver prompt $[P_c,\, \mathrm{kv}(d_1),\ldots,\mathrm{kv}(d_m),\, Q]$, where $P_c$ is the receiver's instruction prefix and $Q$ the question. Every block starts at position zero, so their positions overlap.
For a weight change we keep the architecture, tokenizer, receiver, text,
samples, and scoring fixed: matching tensor shapes are guaranteed,
matching KV values are not. Direct reuse, without any repair, is tested in
all four cells. The repair methods run in the two different-context cells:
the main results use the cell with the same weights, and
\S\ref{sec:crosscheckpoint} the cell with a weight change.

\subsection{Metrics}
\label{sec:pgr}
\label{sec:frontiers}
We score answers with token-level F1, LongBench's official metric for
these subsets \citep{longbench2024}. Their reference answers are short,
so F1 needs no judge model and scores every submission the same way.
For each model and subset we fix two
reference scores: the \emph{floor}, where the receiver answers without
the documents, and the \emph{ceiling}, where it recomputes the
complete prompt over the \emph{same text}. Because the ceiling reads the
same text as the method, a loss against it reflects reuse, not missing
information. A method's quality is the share of the floor-to-ceiling gap it
recovers,
\begin{equation}
\mathrm{PGR} \;=\; \frac{S_{\mathrm{method}} - S_{\mathrm{floor}}}
                        {S_{\mathrm{ceiling}} - S_{\mathrm{floor}}},
\label{eq:pgr}
\end{equation}
which is $0$ at the floor, $1$ at the ceiling, may exceed $1$, and may
be negative: a reused cache can be worse than no context. The floor and ceiling are computed once per model and subset, so all
methods share the same denominator. On 8B, the context is worth
$.326/.361/.338$ F1 on the three subsets; Table~\ref{tab:headroom} gives
every model.

Methods spend and save different resources, and there is no agreed way to
convert one into another, so each cost is reported on its own.
\textbf{Compute}:
Compute saved $= 1 - R/D$, where $R$ is the number of document tokens a method recomputes, counted once per layer, and $D$ the number full recomputation computes, both summed over samples.
\textbf{Memory}: Memory saved $= 1 - B_{\mathrm{method}}/B_{\mathrm{full}}$,
measured on the actual tensors (scales, indices, and copies included)
against the uncompressed BF16 cache. \textbf{Latency}: TTFT saved $= 1 - T_{\mathrm{method}}/T_{\mathrm{full}}$, where $T$ is the time to first token (TTFT), measured per request with the cache already available, with the method and full recomputation on the same backend. Building the caches is a one-time cost outside this measure; how soon it pays off depends on how often a workload reads the cache
(Appendix~\ref{app:amortization}).

\textbf{Score}:\label{sec:leaderboard} The leaderboard score combines quality and speed and ranks all methods in one list. As in RouterArena \citep{routerarena2025}, the score is a weighted sum, $S = \tfrac12 Q + \tfrac12 V$. $Q$ is the mean PGR over the five models and three subsets, and $V$ is the TTFT saved, averaged over the same models and subsets, each relative to the model's own full recomputation. Memory saved is compared separately (Table~\ref{tab:memory}). Position alignment, the reference, is not scored. Users can change the weight on the leaderboard, and
Appendix~\ref{app:score} lists three presets.
A new method enters through the three calls of Figure~\ref{fig:overview};
Appendix~\ref{sec:framework} describes the submission steps.

\section{Experiments}
\label{sec:results}

After the methods and settings (\S\ref{sec:exp-settings}), we rank the
methods after a context change and break the ranking into quality and
costs (\S\ref{sec:leaderboard-results}--\S\ref{sec:frontier-results}).
We then look at the effect of the receiver model
(\S\ref{sec:confirmation}) and of the producer model
(\S\ref{sec:crosscheckpoint}).

\begin{figure}[t]
\begin{minipage}[t]{0.465\linewidth}
\vspace{0pt}
\centering
\includegraphics[width=\linewidth]{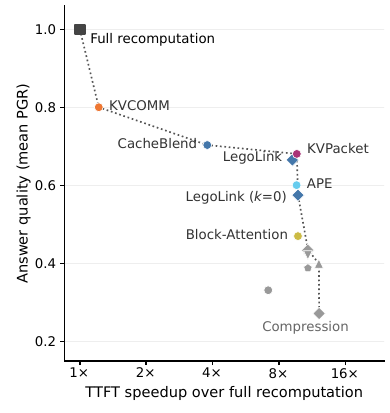}
\caption{With the caches already available, most methods cut time to first token about tenfold relative to full recomputation and differ mainly in quality. CacheBlend gives up part of that speed for quality, and compression (gray) keeps less quality at a similar speed.
Quality is the mean PGR over five models and three QA subsets; speed is
the TTFT speedup over full recomputation, from the share of TTFT saved on
the same models and subsets. Dotted line: methods that no other method beats on both quality and speed.}
\label{fig:pareto}
\end{minipage}\hfill
\begin{minipage}[t]{0.5\linewidth}
\vspace{0pt}
\expandafter\def\csname @captype\endcsname{table}
\caption{The KVShareArena leaderboard: repair methods take the top six places, and compression keeps less quality at a similar speed. Score $=\tfrac12$ quality $+\tfrac12$ speed for every method. Twelve of the 14 entries have speed scores of $.86$--$.92$, so quality decides most of the order. Quality is as in Figure~\ref{fig:pareto}; speed is the share of TTFT saved, averaged over the same models and subsets. KV memory is compared separately (Table~\ref{tab:memory}). LegoLink recomputes the first $k$ tokens of each
block after the first ($k{=}2$ unless marked).
$^{*}$Recomputes the questions whose documents it has not seen.}
\label{tab:leaderboard}
\vspace{4pt}
\centering
\scriptsize
\setlength{\tabcolsep}{4pt}
\renewcommand{\arraystretch}{1.1}
\begin{tabular}{@{}rlrrr@{}}
\toprule
\# & Method & Quality & Speed & Score \\
\midrule
1 & \cls{7}~KVPacket & .681 & .896 & .789 \\
2 & \cls{1}~LegoLink & .666 & .891 & .778 \\
3 & \cls{2}~APE & .601 & .896 & .748 \\
4 & \cls{1}~LegoLink ($k{=}0$) & .575 & .898 & .736 \\
5 & \cls{1}~CacheBlend & .704 & .736 & .720 \\
6 & \cls{6}~Block-Attention & .470 & .897 & .684 \\
7 & \cls{9}~SnapKV ($r{=}.25$) & .435 & .908 & .671 \\
8 & \cls{9}~Knorm ($r{=}.25$) & .434 & .907 & .671 \\
9 & \cls{9}~TOVA ($r{=}.25$) & .421 & .908 & .664 \\
10 & \cls{9}~SnapKV ($r{=}.5$) & .399 & .918 & .659 \\
11 & \cls{9}~StreamingLLM ($r{=}.25$) & .388 & .908 & .648 \\
12 & \cls{9}~Quant.\ 4-bit & .331 & .860 & .596 \\
13 & \cls{9}~Knorm ($r{=}.5$) & .272 & .918 & .595 \\
14 & \cls{4}~KVCOMM$^{*}$ & .801 & .179 & .490 \\
\bottomrule
\end{tabular}

\end{minipage}
\end{figure}

\subsection{Methods and settings}
\label{sec:exp-settings}
\label{sec:methods}
The main evaluation compares six method classes against position alignment (\cls{3}, the reference), marked by the same glyphs in all tables: \cls{1}~selective recomputation, \cls{2}~attention calibration, \cls{4}~anchor-delta reuse (correcting a cache with offsets stored from earlier requests), \cls{6}~fine-tuned repair, \cls{7}~trained soft-token adapter (a few trained tokens wrapped around each cached block; the model itself is unchanged), and \cls{9}~compression under reuse. Table~\ref{tab:matrix} (appendix) lists every method, including two further classes (\cls{5}, \cls{8}).
Each method uses the settings of its paper or official implementation.
Where a method has a budget to choose, we run LegoLink at $k{=}2$ and
$k{=}0$ recomputed tokens per block and the eviction methods at $r{=}.25$, the
share of cached tokens removed, with SnapKV and Knorm also at $r{=}.5$.
Every comparison is paired: both sides answer the same questions.
We write $n/5$ for the number of models on which a method
significantly beats position alignment (paired bootstrap, 95\%). Appendix~\ref{app:settings}
lists the run settings.

\subsection{Overall ranking}
\label{sec:leaderboard-results}
Repair methods take the top six places on the leaderboard, and compression
keeps less quality at a similar speed (Table~\ref{tab:leaderboard}). Most
methods cut time to first token about tenfold, so quality decides most of the order (Figure~\ref{fig:pareto}). KVPacket and LegoLink lead. CacheBlend and KVCOMM keep more quality but give up speed for it, so they rank fifth and last. With a quality weight of $0.7$ or $0.3$ instead of $0.5$ (Appendix~\ref{app:score}), KVPacket and LegoLink
stay first and second, and repair methods keep the top five places. Per-dataset and per-model results are
in Appendix~\ref{app:boards}.

\subsection{Answer quality}
\label{sec:re-board}
\begin{table}[t]
\caption{\textbf{Repair beats position alignment on single-document and
multi-hop QA, and the leading repairs outperform compression.} Mean PGR
over the five models, each against its own floor and ceiling.
Superscripts count the models on which the method is significantly above
position alignment (paired bootstrap, 95\%); \textbf{bold} = above on at
least three models. Bold entries come from five method classes.}
\label{tab:re-board}
\begin{center}
\small
\setlength{\tabcolsep}{4pt}
\begin{tabular}{@{}lrrr@{}}
\toprule
Method & Sci.-paper QA & Single-doc QA & Multi-hop QA \\
\midrule
\cls{3}~Position alignment & .630 & .550 & .209 \\
\midrule
\cls{1}~CacheBlend (15\% recompute) & \textbf{.868}$^{\uparrow5}$ & \textbf{.807}$^{\uparrow5}$ & \textbf{.436}$^{\uparrow4}$ \\
\cls{1}~LegoLink ($k{=}2$, $\approx$0.4\% recompute) & \textbf{.802}$^{\uparrow4}$ & \textbf{.732}$^{\uparrow4}$ & \textbf{.463}$^{\uparrow4}$ \\
\cls{1}~LegoLink ($k{=}0$, 0\% recompute) & .718$^{\uparrow2}$ & \textbf{.642}$^{\uparrow3}$ & \textbf{.366}$^{\uparrow3}$ \\
\cls{2}~APE & .605 & .702$^{\uparrow2}$ & \textbf{.496}$^{\uparrow5}$ \\
\cls{4}~KVCOMM & .546$^{\uparrow1}$ & \textbf{.857}$^{\uparrow5}$ & \textbf{.999}$^{\uparrow5}$ \\
\cls{6}~Block-Attention & .556 & .551 & \textbf{.304}$^{\uparrow4}$ \\
\cls{7}~KVPacket (0\% recompute) & \textbf{.788}$^{\uparrow5}$ & \textbf{.737}$^{\uparrow4}$ & \textbf{.518}$^{\uparrow5}$ \\
\cls{9}~Quant.\ 4-bit & .543 & .483 & $-$.032 \\
\cls{9}~SnapKV ($r{=}.25$) & .600 & .494 & .212$^{\uparrow1}$ \\
\cls{9}~SnapKV ($r{=}.5$) & .552 & .431 & .215 \\
\cls{9}~Knorm ($r{=}.25$) & .560 & .470 & .271$^{\uparrow1}$ \\
\cls{9}~Knorm ($r{=}.5$) & .387 & .265 & .163 \\
\cls{9}~TOVA ($r{=}.25$) & .586 & .482 & .197 \\
\cls{9}~StreamingLLM ($r{=}.25$) & .628 & .504 & .032 \\
\bottomrule
\end{tabular}

\end{center}
\end{table}
Repair improves answer quality on all three QA subsets, and the largest
unrecovered gap remains on multi-hop QA.

\begin{wraptable}{R}{0.43\textwidth}
\vspace{-10pt}
\caption{\textbf{Compression saves $25$--$72\%$ of KV memory but keeps
less quality than every repair method.} Means over five models and three
QA subsets; quality is PGR.}
\label{tab:memory}
\vspace{3pt}
\centering
\scriptsize
\setlength{\tabcolsep}{2pt}
\begin{tabular}{@{}lrr@{}}
\toprule
Method & Memory saved & Quality \\
\midrule
\cls{9}~Quant.\ 4-bit & .719 & .331 \\
\cls{9}~SnapKV ($r{=}.5$) & .500 & .399 \\
\cls{9}~Knorm ($r{=}.5$) & .500 & .272 \\
\cls{9}~SnapKV ($r{=}.25$) & .250 & .435 \\
\cls{9}~Knorm ($r{=}.25$) & .250 & .434 \\
\cls{9}~TOVA ($r{=}.25$) & .250 & .421 \\
\cls{9}~StreamingLLM ($r{=}.25$) & .250 & .388 \\
\bottomrule
\end{tabular}

\vspace{-6pt}
\end{wraptable}

The comparisons below average the
five models, each scored against its own floor and ceiling
(Table~\ref{tab:re-board}; Figure~\ref{fig:boards} in the appendix):
\textbf{1)}~Position alignment alone recovers $.63$, $.55$, and $.21$ of the score
gap on scientific-paper, single-document, and multi-hop QA. It leaves
room for repair on all three subsets.
\textbf{2)}~Selective recomputation \cls{1} improves all three subsets.
CacheBlend beats position alignment on $5/5$, $5/5$, and $4/5$ models,
respectively; LegoLink does so on $4/5$ for each. Attention
calibration \cls{2} improves multi-hop QA on all five models. These
repairs lift the multi-hop mean to $.44$--$.50$.
\textbf{3)}~KVPacket \cls{7} improves all three subsets without
recomputing any document token. It reaches mean PGR $.79/.74/.52$ and
beats position alignment on $5/5$, $4/5$, and $5/5$ models, respectively.
\textbf{4)}~Compression \cls{9} keeps less quality than the leading repairs (Table~\ref{tab:memory}); its best multi-hop mean is
$.27$.

\subsection{Costs: compute, memory, and latency}
\label{sec:frontier-results}
Compute and latency separate methods by how many tokens they recompute;
memory separates compression from repair. Each saving is measured against the same
model's full recomputation and then averaged over the five models.

\textbf{Compute.} Most methods recompute none of the cached document tokens (Figure~\ref{fig:boards}, appendix). Their main difference is answer quality. Among methods that do
recompute, CacheBlend, run at a $15\%$ recomputation budget, spends about $17\%$ of the document computation of full recomputation and LegoLink with $k{=}2$ spends under $0.5\%$. Standard prefix caching must recompute all document tokens in this setting and therefore saves no document computation.

\textbf{Memory.} Only compression saves KV bytes, at a cost to quality (Table~\ref{tab:memory}). The repair methods keep the full cache. KVPacket's
soft tokens add about $3\%$.

\textbf{Latency.} Methods that recompute more tokens save less TTFT (Table~\ref{tab:leaderboard}). Repairs that recompute no document tokens save $.81$--$.94$ of TTFT across models and
datasets. CacheBlend saves $.66$--$.78$.

Figures~\ref{fig:frontier-runtime} and~\ref{fig:frontier-memory} in the
appendix show the dataset-level results on Qwen3-8B.

\subsection{Effect of the receiver model}
\label{sec:confirmation}
\label{sec:crossmodel}
Which repairs help depends on the receiver model. No method leads on every model: on multi-hop QA, LegoLink and APE reach $.50$ and $.49$ PGR on 8B,
KVPacket reaches $.54$ on 4B, $.56$ on 14B, and $.61$ on Llama, and
CacheBlend reaches $.57$ on 30B (appendix). Models also differ in how well
they read separately cached blocks: position alignment alone recovers
$.03$ of the multi-hop gap on 8B, compared with $.32$ on 4B, $.19$ on 14B,
and $.41$ on Llama (Table~\ref{tab:headroom}). On Llama,
KVPacket, APE, and Block-Attention significantly improve the multi-hop
score, while CacheBlend and LegoLink do not. Choosing a method
therefore needs results on more than one model. Some gains still hold on
every model: CacheBlend gains at least $.20$ PGR over position alignment
on single-document QA, and KVPacket and APE gain at least $.20$ and
$.11$, respectively, on multi-hop QA (Table~\ref{tab:robustness},
appendix).

\begin{table}[t]
\caption{\textbf{Models differ in how much position alignment leaves to
repair.} Worth = ceiling$-$floor in F1; left = the share of that gap
still missing after position alignment.}
\label{tab:headroom}
\begin{center}
\scriptsize
\setlength{\tabcolsep}{3.5pt}
\begin{tabular}{@{}lrrrrrr@{}}
\toprule
Board & \multicolumn{2}{c}{Sci.-paper QA} & \multicolumn{2}{c}{Single-doc QA} & \multicolumn{2}{c}{Multi-hop QA} \\
 & worth & left & worth & left & worth & left \\
\midrule
Qwen3-8B & .326 & .432 & .361 & .597 & .338 & .970 \\
Qwen3-4B & .322 & .496 & .357 & .592 & .353 & .684 \\
Llama-3.1-8B-Instruct & .319 & .249 & .474 & .352 & .289 & .591 \\
Qwen3-14B & .366 & .349 & .327 & .417 & .327 & .806 \\
Qwen3-30B-A3B & .304 & .324 & .364 & .289 & .316 & .904 \\
\bottomrule
\end{tabular}

\end{center}
\end{table}

\subsection{Effect of the producer model}
\label{sec:crosscheckpoint}
Most repairs keep their quality when another version of the model writes
the cache, while KVPacket's trained adapter needs retraining. The producer
is either a sibling post-trained from the same base, as when a service updates its
post-trained model, or
the receiver's own base model before post-training, the largest weight
change within one family; Table~\ref{tab:pairs} (appendix) lists the six
pairs. These tests show how existing reuse methods respond to a new producer. DroidSpeak, a system designed for weight changes, has not released its code (Appendix~\ref{app:matrix}). Without repair
(Table~\ref{tab:factorial-results}, appendix), every cross-context score
is lower when a different producer wrote the cache, but the drop size
depends on task and producer both.

Re-running every repair method with a same-family producer on all five
models shows which repairs depend on the producer
(Table~\ref{tab:robustness}, appendix). Most lose less than $.02$ F1 on
average and at most $.04$ in any test (Appendix~\ref{sec:robustness});
CacheBlend has no significant drop in any test. Two repairs change more. KVPacket, whose adapter is trained on the receiver's own caches,
falls significantly below its same-producer score in $12$ of the $18$
tests (six producer pairs $\times$ three QA subsets; worst $-.146$ F1).
LegoLink loses up to $.151$ F1 without recomputation ($k{=}0$), while
with $k{=}2$ it stays within $.04$.

Of the two trained repairs, KVPacket changes with the producer and
Block-Attention does not (less than $.01$ F1 on average).
Does the adapter need a new design, or only retraining? We retrained the adapter with its published recipe and budget on caches from each new producer and tested it on the six tests with 8B as the receiver (two producers $\times$ three subsets; appendix). We compare it with the original adapter and with a cross-trained adapter, trained on the other producer. Retraining on
the actual producer substantially recovers the lost quality, by up to $.25$ F1 where the original adapter had lost the most, and we
detect no remaining difference from the same-model baseline in any test. The cross-trained adapter recovers only part of that loss and
stays below the baseline in every test, so the adapter must be trained on the producer it will read. This retraining closes the gap at about $20$ GPU-hours per producer.

A cache written by a checkpoint from a different pretraining run
lowers every method, by up to $.32$ F1
(Table~\ref{tab:diff-pretrain}, appendix); reuse across such checkpoints
is still open.

\section{Related Work}
\label{sec:related}

\paragraph{Four forms of KV reuse.}
We group work on KV reuse into four forms. \textbf{(i) Prefix caching}
\citep{radixattention2024} reuses KV only under an exact match. It is
lossless and aimed at throughput. In our setting the preceding text
always changes, so it recomputes every document (Appendix~\ref{sec:ecosystem}). \textbf{(ii) Reuse after the preceding text or the model changes, with
repair}, the setting this paper evaluates: selective recomputation
\citep{cacheblend2025,epic2025,droidspeak2025},
attention calibration \citep{ape2025}, and anchor-delta reuse
\citep{kvcomm2025}. \textbf{(iii)
Cache storage}: caches moved across the memory hierarchy
\citep{cachedattention2024}, a separate problem from repair. \textbf{(iv) Input
rewriting}: reordering contexts to create common prefixes
\citep{contextpilot2026}; it changes the input rather than repairing the
cache.

\paragraph{Repair methods and their own evaluations.} Repair methods are evaluated in their own papers, each on its own tasks and cost measures. The earliest form of (ii) is modular reuse with position re-encoding \citep{promptcache2024}. RAG-serving methods report F1 changes and TTFT on chunked QA. Fine-tuned repair
\citep{turborag2024,kvlink2025,blockattention2024} retrains the model; we include it and report its training cost. The newest method is a trained soft-token adapter \citep{kvpacket2026}. RelayCaching
\citep{relaycaching2026}, which repairs the KV an agent produced while writing its output, appears in our supplementary Agent Reports
experiment (Appendix~\ref{app:agent-reports}). KVCOMM \citep{kvcomm2025} corrects
caches with offsets stored from earlier requests. Latent communication \citep{latentmas2026} changes what agents exchange, hidden states instead of text,
and does not repair a cache. Cross-architecture fusion
\citep{c2c2025} targets models of different architectures.

\paragraph{Benchmarks and evaluations.}
SCBench \citep{scbench2025} evaluates
long-context methods through shared-prefix, multi-turn, and multi-request
reuse. Each cache is written and read by the same model.
KVShareArena instead benchmarks reuse after the preceding text or
model weights change. It compares the retained answer quality and repair
cost across methods.

Cross-context reuse has also received focused empirical study. \citet{elevenmethodstudy2026} compare eleven methods for cross-context
reuse on answer quality but do not measure serving cost.
\citet{kvreusefails2026} study reuse failures in multi-agent
evaluation. DroidSpeak \citep{droidspeak2025} develops a cross-checkpoint
sharing system and evaluates its own method. These studies motivate the problem but do not offer a shared benchmark that compares quality and cost when the context or the checkpoint changes.

RouterArena \citep{routerarena2025} provides a shared evaluation protocol
and an open leaderboard for LLM routers. KVShareArena adopts this open
comparison approach for KV reuse. HELMET and MemoryBench
\citep{helmet2025,memorybench2025} benchmark long-context models and
memory in LLM systems; LongBench \citep{longbench2024} supplies our QA tasks.

\section{Conclusion}
\label{sec:conclusion}
KVShareArena compares KV-cache reuse methods under the same conditions
when the preceding text, the producer, or both change, and reports
quality, compute, memory, and latency separately. The experiments yield two findings. First, both the quality loss from reuse and which repairs help depend on the receiver, even between two 8B models.
Second, most repairs keep their quality when another version of the model wrote
the cache, but a trained repair adapter can need retraining for each new
producer. The experiments also show that compression saves memory but recovers
less quality than the leading repairs.
These results point to repairs that also shrink the cache and to repairs
that adapt to new producers at lower cost. Reuse across models from
different pretraining runs remains open. KVShareArena's common interface
and leaderboard let new methods join the same comparison.

\clearpage
\section*{Reproducibility Statement}
The public release provides an installable evaluation package,
the fixed question sets, reference scores for the five receiver models,
an example driver for position alignment, and an interactive leaderboard (\repourl; data at \dataseturl; leaderboard at
\leaderboardurl).{}
The package re-scores per-sample outputs on CPU and computes F1 and
PGR. Section~\ref{sec:setup} and Appendices~\ref{app:matrix},
\ref{app:data}, and~\ref{app:settings} give the models, method settings,
datasets, and run settings;
Appendix~\ref{app:platform} describes the release and submission steps.

\section*{Ethics Statement}
This work evaluates methods for reusing KV caches on public
question-answering data (LongBench, Qasper, HotpotQA, and Wikipedia text)
with publicly released model checkpoints. It involves no human subjects
and collects no personal data. Sharing caches across requests, users, or
models raises privacy and security questions, which we discuss in
Appendix~\ref{sec:limitations}. The released question sets keep the
licenses of their sources.

\section*{AI Use Statement}
AI assistance was used for engineering, writing, and auxiliary data
analysis under author review. All experimental decisions, claims, and
final text are the authors' responsibility.

\bibliography{refs}
\bibliographystyle{iclr2027_conference}

\appendix
\clearpage
\etocdepthtag.toc{mtappendix}
\etocsettagdepth{mtmain}{none}
\etocsettagdepth{mtappendix}{section}
\etocsettocstyle{\section*{Appendix Overview}}{}
\etocsetstyle{section}{\par\smallskip}{\noindent}
  {\makebox[1.6em][l]{\etocnumber}\etocname\dotfill\etocpage\par}{\medskip}
\tableofcontents
\section{Limitations}
\label{sec:limitations}
Sharing caches across requests, users, or models also raises security and
privacy questions. A cache is computed from its text, so it may reveal
that text to anyone who can read or probe it. A cache hit is faster than
a miss, so response times alone can hint at what another user has sent.
A cache written by one party and read by another can also be altered, and
the reader cannot easily verify it. Deployments therefore need to decide
which caches may be shared, with whom, and for how long. These questions
are separate from the quality and cost trade-offs we measure, and we
leave them to future work.

\section{Detailed Leaderboard Results}
\label{app:boards}
\begin{figure}[ht]
\centering
\includegraphics[width=\linewidth]{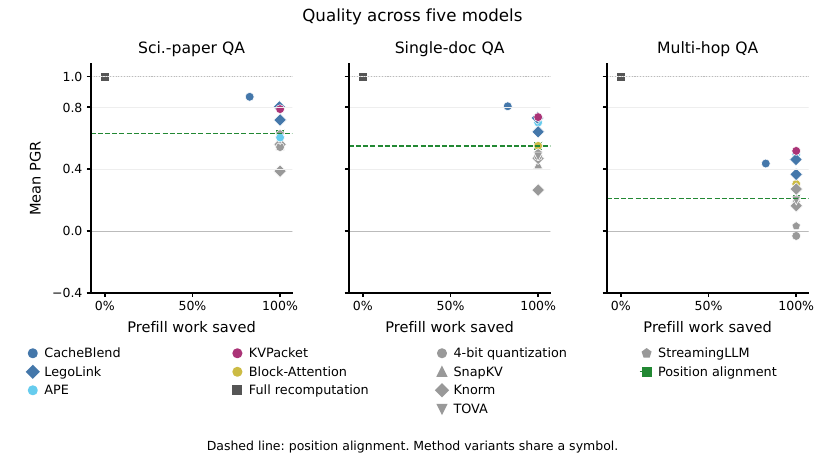}
\caption{Dataset-level quality and recomputation cost. Each panel is one
QA subset of the main evaluation. PGR is averaged over the five models, each against its own floor and
ceiling. The horizontal axis shows Compute saved on the Qwen3-8B board. Higher and farther right is better. The dashed
line marks position alignment. Configurations of the same method
share a symbol. Table~\ref{tab:re-board} lists scores and paired
comparison counts.}
\label{fig:boards}
\end{figure}

Table~\ref{tab:re-board} gives the numbers behind Figure~\ref{fig:boards}: each method's mean PGR over the five models and the number of models on which it is significantly above position alignment. One table per model follows; we call each model's results a \emph{board}.

\begin{table}[h]
\caption{Qwen3-8B board (reference board). PGR (F1);
$^{\uparrow}$~= significantly above position alignment (paired, 95\%). KVCOMM keeps the offsets it stores from earlier questions and recomputes the questions whose documents it has not seen
(Table~\ref{tab:leaderboard}).}
\label{tab:board-8B}
\begin{center}
\scriptsize
\setlength{\tabcolsep}{2.5pt}
\begin{tabular}{@{}lrrr@{}}
\toprule
Method & Sci.-paper QA & Single-doc QA & Multi-hop QA \\
\midrule
\cls{3}~Position alignment & .568 & .403 & .030 \\
\midrule
\cls{1}~CacheBlend (15\% recompute) & .857$^{\uparrow}$ & .748$^{\uparrow}$ & .440$^{\uparrow}$ \\
\cls{1}~LegoLink ($k{=}2$, $\approx$0.4\% recompute) & .853$^{\uparrow}$ & .674$^{\uparrow}$ & .496$^{\uparrow}$ \\
\cls{1}~LegoLink ($k{=}0$, 0\% recompute) & .569 & .489 & .397$^{\uparrow}$ \\
\cls{2}~APE & .549 & .633$^{\uparrow}$ & .492$^{\uparrow}$ \\
\cls{4}~KVCOMM & .539 & .847$^{\uparrow}$ & 1.001$^{\uparrow}$ \\
\cls{6}~Block-Attention & .489 & .471 & .124$^{\uparrow}$ \\
\cls{7}~KVPacket (0\% recompute) & .750$^{\uparrow}$ & .645$^{\uparrow}$ & .469$^{\uparrow}$ \\
\cls{9}~Quant.\ 4-bit & .388 & .382 & $-$.115 \\
\cls{9}~SnapKV ($r{=}.25$) & .541 & .352 & .026 \\
\cls{9}~SnapKV ($r{=}.5$) & .499 & .301 & .041 \\
\cls{9}~Knorm ($r{=}.25$) & .512 & .425 & .071 \\
\cls{9}~Knorm ($r{=}.5$) & .338 & .215 & $-$.075 \\
\cls{9}~TOVA ($r{=}.25$) & .533 & .359 & .016 \\
\cls{9}~StreamingLLM ($r{=}.25$) & .599 & .398 & $-$.054 \\
\bottomrule
\end{tabular}

\end{center}
\end{table}

\begin{table}[h]
\caption{Qwen3-4B board. PGR (F1);
$^{\uparrow}$~= significantly above position alignment (paired, 95\%).}
\label{tab:board-4B}
\begin{center}
\footnotesize
\setlength{\tabcolsep}{3pt}
\begin{tabular}{@{}lrrr@{}}
\toprule
Method & Sci.-paper QA & Single-doc QA & Multi-hop QA \\
\midrule
\cls{3}~Position alignment & .504 & .408 & .316 \\
\midrule
\cls{1}~CacheBlend (15\% recompute) & .794$^{\uparrow}$ & .685$^{\uparrow}$ & .453$^{\uparrow}$ \\
\cls{1}~LegoLink ($k{=}2$, $\approx$0.4\% recompute) & .802$^{\uparrow}$ & .685$^{\uparrow}$ & .505$^{\uparrow}$ \\
\cls{1}~LegoLink ($k{=}0$, 0\% recompute) & .763$^{\uparrow}$ & .554$^{\uparrow}$ & .288 \\
\cls{2}~APE & .413 & .525 & .545$^{\uparrow}$ \\
\cls{4}~KVCOMM & .576$^{\uparrow}$ & .851$^{\uparrow}$ & .998$^{\uparrow}$ \\
\cls{6}~Block-Attention & .423 & .479 & .397$^{\uparrow}$ \\
\cls{7}~KVPacket (0\% recompute) & .798$^{\uparrow}$ & .658$^{\uparrow}$ & .543$^{\uparrow}$ \\
\cls{9}~Quant.\ 4-bit & .307 & .268 & $-$.390 \\
\cls{9}~SnapKV ($r{=}.25$) & .470 & .303 & .366$^{\uparrow}$ \\
\cls{9}~SnapKV ($r{=}.5$) & .450 & .286 & .362 \\
\cls{9}~Knorm ($r{=}.25$) & .468 & .272 & .356 \\
\cls{9}~Knorm ($r{=}.5$) & .281 & .125 & .174 \\
\cls{9}~TOVA ($r{=}.25$) & .453 & .331 & .322 \\
\cls{9}~StreamingLLM ($r{=}.25$) & .484 & .401 & .198 \\
\bottomrule
\end{tabular}

\end{center}
\end{table}

\begin{table}[h]
\caption{Llama-3.1-8B-Instruct board. PGR (F1);
$^{\uparrow}$~= significantly above position alignment (paired, 95\%).}
\label{tab:board-Llama}
\begin{center}
\footnotesize
\setlength{\tabcolsep}{3pt}
\begin{tabular}{@{}lrrr@{}}
\toprule
Method & Sci.-paper QA & Single-doc QA & Multi-hop QA \\
\midrule
\cls{3}~Position alignment & .751 & .648 & .409 \\
\midrule
\cls{1}~CacheBlend (15\% recompute) & .926$^{\uparrow}$ & .850$^{\uparrow}$ & .332 \\
\cls{1}~LegoLink ($k{=}2$, $\approx$0.4\% recompute) & .753 & .639 & .394 \\
\cls{1}~LegoLink ($k{=}0$, 0\% recompute) & .812 & .775$^{\uparrow}$ & .418 \\
\cls{2}~APE & .820 & .870$^{\uparrow}$ & .519$^{\uparrow}$ \\
\cls{4}~KVCOMM & .461 & .840$^{\uparrow}$ & .997$^{\uparrow}$ \\
\cls{6}~Block-Attention & .644 & .560 & .552$^{\uparrow}$ \\
\cls{7}~KVPacket (0\% recompute) & .863$^{\uparrow}$ & .825$^{\uparrow}$ & .611$^{\uparrow}$ \\
\cls{9}~Quant.\ 4-bit & .744 & .616 & .380 \\
\cls{9}~SnapKV ($r{=}.25$) & .744 & .601 & .395 \\
\cls{9}~SnapKV ($r{=}.5$) & .667 & .538 & .401 \\
\cls{9}~Knorm ($r{=}.25$) & .615 & .480 & .452 \\
\cls{9}~Knorm ($r{=}.5$) & .456 & .326 & .463 \\
\cls{9}~TOVA ($r{=}.25$) & .701 & .591 & .419 \\
\cls{9}~StreamingLLM ($r{=}.25$) & .766 & .632 & .102 \\
\bottomrule
\end{tabular}

\end{center}
\end{table}

\begin{table}[h]
\caption{Qwen3-14B board. PGR (F1);
$^{\uparrow}$~= significantly above position alignment (paired, 95\%).}
\label{tab:board-14B}
\begin{center}
\footnotesize
\setlength{\tabcolsep}{3pt}
\begin{tabular}{@{}lrrr@{}}
\toprule
Method & Sci.-paper QA & Single-doc QA & Multi-hop QA \\
\midrule
\cls{3}~Position alignment & .651 & .583 & .194 \\
\midrule
\cls{1}~CacheBlend (15\% recompute) & .858$^{\uparrow}$ & .811$^{\uparrow}$ & .386$^{\uparrow}$ \\
\cls{1}~LegoLink ($k{=}2$, $\approx$0.4\% recompute) & .853$^{\uparrow}$ & .818$^{\uparrow}$ & .434$^{\uparrow}$ \\
\cls{1}~LegoLink ($k{=}0$, 0\% recompute) & .682 & .569 & .274$^{\uparrow}$ \\
\cls{2}~APE & .607 & .742 & .504$^{\uparrow}$ \\
\cls{4}~KVCOMM & .609 & .862$^{\uparrow}$ & .999$^{\uparrow}$ \\
\cls{6}~Block-Attention & .617 & .542 & .295$^{\uparrow}$ \\
\cls{7}~KVPacket (0\% recompute) & .746$^{\uparrow}$ & .736$^{\uparrow}$ & .555$^{\uparrow}$ \\
\cls{9}~Quant.\ 4-bit & .620 & .556 & .125 \\
\cls{9}~SnapKV ($r{=}.25$) & .619 & .518 & .185 \\
\cls{9}~SnapKV ($r{=}.5$) & .554 & .441 & .153 \\
\cls{9}~Knorm ($r{=}.25$) & .683 & .611 & .320$^{\uparrow}$ \\
\cls{9}~Knorm ($r{=}.5$) & .431 & .290 & .249 \\
\cls{9}~TOVA ($r{=}.25$) & .599 & .509 & .152 \\
\cls{9}~StreamingLLM ($r{=}.25$) & .621 & .469 & $-$.007 \\
\bottomrule
\end{tabular}

\end{center}
\end{table}

\begin{table}[h]
\caption{Qwen3-30B-A3B-Instruct-2507 board (mixture of experts). PGR (F1);
$^{\uparrow}$~= significantly above position alignment (paired, 95\%).}
\label{tab:board-30B}
\begin{center}
\footnotesize
\setlength{\tabcolsep}{3pt}
\begin{tabular}{@{}lrrr@{}}
\toprule
Method & Sci.-paper QA & Single-doc QA & Multi-hop QA \\
\midrule
\cls{3}~Position alignment & .676 & .711 & .096 \\
\midrule
\cls{1}~CacheBlend (15\% recompute) & .905$^{\uparrow}$ & .941$^{\uparrow}$ & .572$^{\uparrow}$ \\
\cls{1}~LegoLink ($k{=}2$, $\approx$0.4\% recompute) & .751$^{\uparrow}$ & .843$^{\uparrow}$ & .485$^{\uparrow}$ \\
\cls{1}~LegoLink ($k{=}0$, 0\% recompute) & .765$^{\uparrow}$ & .821$^{\uparrow}$ & .454$^{\uparrow}$ \\
\cls{2}~APE & .634 & .738 & .421$^{\uparrow}$ \\
\cls{4}~KVCOMM & .546 & .883$^{\uparrow}$ & 1.002$^{\uparrow}$ \\
\cls{6}~Block-Attention & .606 & .702 & .153 \\
\cls{7}~KVPacket (0\% recompute) & .787$^{\uparrow}$ & .820 & .412$^{\uparrow}$ \\
\cls{9}~Quant.\ 4-bit & .656 & .595 & $-$.160 \\
\cls{9}~SnapKV ($r{=}.25$) & .627 & .693 & .086 \\
\cls{9}~SnapKV ($r{=}.5$) & .589 & .590 & .119 \\
\cls{9}~Knorm ($r{=}.25$) & .524 & .563 & .156 \\
\cls{9}~Knorm ($r{=}.5$) & .430 & .369 & .004 \\
\cls{9}~TOVA ($r{=}.25$) & .643 & .618 & .074 \\
\cls{9}~StreamingLLM ($r{=}.25$) & .669 & .622 & $-$.080 \\
\bottomrule
\end{tabular}

\end{center}
\end{table}

\clearpage
\section{Supplementary Experiment: Agent Reports}
\label{app:agent-reports}
\label{sec:ar-board}
This experiment asks whether repair methods also help when the shared
text consists of specialist reports rather than raw document chunks.
It uses the same HotpotQA questions as the main evaluation, which we
call Retrieved Evidence here. Results are
reported separately and do not enter the main Retrieved Evidence means.
On each board, the producers and the receiver (the agent that reads the reports and answers) use the same model.

\subsection{Report construction and evaluation}
Specialist agents each read a disjoint part of the source material and
write a report without seeing the question. Each board's reports are written once and then fixed, so every method receives the same report text.

\begin{figure}[t]
\centering
\begin{tikzpicture}[
  font=\small,
  box/.style={draw,rounded corners=1pt,inner sep=2.5pt,font=\scriptsize,align=center},
  chunk/.style={box,minimum width=0.62cm,minimum height=0.38cm,text=white},
  agent/.style={box,fill=gray!12,minimum width=1.2cm},
  arr/.style={->,>=stealth,gray!70,thick},
  lane/.style={font=\scriptsize\bfseries,anchor=west},
  note/.style={font=\tiny,align=center,text=gray!40!black}]

\node[lane] at (0,1.75) {Main evaluation: Retrieved Evidence};
\node[note] at (2.55,1.45) {encoded alone, cached offline};
\node[box,fill=gray!12,align=center] (docs) at (0.75,0.55) {corpus\\ docs};
\node[chunk,fill=blkA] (c1) at (2.55,1.05) {$d_1$};
\node[chunk,fill=blkB] (c2) at (2.55,0.55) {$d_2$};
\node[chunk,fill=blkC] (c3) at (2.55,0.05) {$d_3$};
\node[box,fill=gray!5,minimum width=3.1cm] (asm1) at (7.0,0.55)
  {$[P_c,\,\mathrm{kv}(d_1),\,\mathrm{kv}(d_2),\,\mathrm{kv}(d_3),\,Q]$};
\node[agent] (llm1) at (10.0,0.55) {receiver};
\node[font=\scriptsize] (ans1) at (11.5,0.55) {answer};
\draw[arr] (docs) -- node[above,font=\tiny,text=gray]{chunk} (2.15,0.55);
\draw[arr] (2.95,0.55) -- node[below,font=\tiny,text=gray]{assemble at query time} (asm1);
\draw[arr] (asm1) -- (llm1);
\draw[arr] (llm1) -- (ans1);
\node[note,anchor=west] at (5.45,-0.15) {ceiling: full recomputation over the \emph{same blocks}};

\node[lane] at (0,-1.05) {Supplementary experiment: Agent Reports};
\node[box,fill=gray!12,align=center] (mat) at (0.75,-2.15) {source\\ material (split)};
\node[agent] (a1) at (2.75,-1.65) {producer 1};
\node[agent] (a2) at (2.75,-2.15) {producer 2};
\node[agent] (a3) at (2.75,-2.65) {producer 3};
\node[chunk,fill=blkA] (r1) at (4.55,-1.65) {$r_1$};
\node[chunk,fill=blkB] (r2) at (4.55,-2.15) {$r_2$};
\node[chunk,fill=blkC] (r3) at (4.55,-2.65) {$r_3$};
\node[box,fill=gray!5,minimum width=3.1cm] (asm2) at (7.6,-2.15)
  {$[P_c,\,\mathrm{kv}(r_1),\,\mathrm{kv}(r_2),\,\mathrm{kv}(r_3),\,Q]$};
\node[agent] (llm2) at (10.6,-2.15) {receiver};
\node[font=\scriptsize] (ans2) at (12.1,-2.15) {answer};
\draw[arr] (mat.east) -- (a1.west);
\draw[arr] (mat.east) -- (a2.west);
\draw[arr] (mat.east) -- (a3.west);
\draw[arr] (a1) -- (r1);
\draw[arr] (a2) -- node[above,font=\tiny,text=gray]{reports} (r2);
\draw[arr] (a3) -- (r3);
\draw[arr] (4.95,-2.15) -- (asm2);
\draw[arr] (asm2) -- (llm2);
\draw[arr] (llm2) -- (ans2);
\node[note] at (2.75,-3.1) {each reads one part, without the question};
\node[note,anchor=west] at (6.05,-2.85) {ceiling: full recomputation over the \emph{same report text}};
\node[note,anchor=west] at (6.05,-3.15) {floor (both): the receiver answers without the shared text};

\draw[decorate,decoration={brace,amplitude=4pt},gray]
  (-0.3,1.9) -- (-0.3,-3.35);
\node[rotate=90,font=\scriptsize,text=gray!40!black,align=center]
  at (-0.65,-0.7) {same operation: assemble\\ independently encoded blocks};
\end{tikzpicture}
\caption{Document assembly in the main evaluation and report assembly
in the supplement. The first uses separately cached document blocks. The second uses fixed reports written without the question. The diagram shows how most methods build the caches; RelayCaching instead reuses the KV produced while each report was written. Each experiment has its own ceiling (\S\ref{sec:pgr}), so scores are not averaged across the two.}
\label{fig:tracks}
\end{figure}
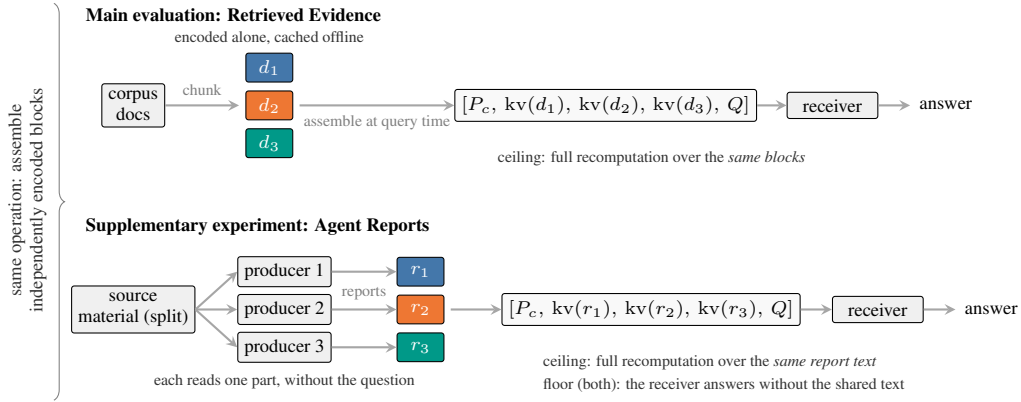

Most methods encode each report on its own and pass its cache to the receiver (Figure~\ref{fig:tracks}). RelayCaching instead reuses the KV produced while each report was written, obtained by replaying the fixed report text, and repairs it. The two procedures can give different KV for the same report.
The benchmark records the cost of cache construction separately from
per-request reuse.

The no-context floor is unchanged. The ceiling is full recomputation over the
\emph{same report text}, not the original documents. Reports can leave out information. A ceiling over the original documents would count that loss too, so the ceiling uses the report text. PGR is therefore not averaged across
Retrieved Evidence and Agent Reports. Answer scoring uses the same token-level F1. Table~\ref{tab:report-headroom} gives the
report-specific context value and the room left for repair.

\begin{table}[t]
\caption{Agent Reports: context value and room for repair, under the same
definitions as Table~\ref{tab:headroom}. Worth = ceiling minus floor;
left = the share of that gap still missing after position alignment.
$^{\ddagger}$~worth less than $.25$ F1.}
\label{tab:report-headroom}
\centering
\small
\begin{tabular}{@{}lrr@{}}
\toprule
Board & worth & left \\
\midrule
Qwen3-8B & .252 & .757 \\
Qwen3-4B & .316 & .490 \\
Llama-3.1-8B-Instruct & .152$^{\ddagger}$ & 1.030 \\
Qwen3-14B & .303 & .425 \\
Qwen3-30B-A3B & .237$^{\ddagger}$ & .796 \\
\bottomrule
\end{tabular}

\end{table}

\subsection{Method coverage}
Table~\ref{tab:matrix} lists which methods run here. RelayCaching repairs the KV an agent produced while writing its report, so it runs on Agent Reports and not on Retrieved Evidence.
KVCOMM is not included here: it corrects caches with offsets stored from earlier requests, and the fixed reports do not repeat.

\subsection{Diagnostic controls}
On Qwen3-8B, the share of the gap that position alignment leaves rises with the number of producers, reaching $.76$ (95\% interval $[.46,1.19]$) when the material is split across all producers. If CacheBlend recomputes randomly chosen tokens instead of the tokens it selects, at the same $15\%$ budget, most of its gain disappears where reuse loses the most.

Every method reads the same fixed reports, written without the question, so score differences come from reuse. These results supplement the Retrieved Evidence evaluation.

\clearpage
\section{When Position Alignment Is Enough}
\label{app:boundary}

\label{sec:findings}

When one agent hands a single cache to another, position alignment alone is enough. In this test, a Qwen3-8B agent reads the documents and writes a note, and a second agent answers from the note's KV cache without the documents (100 questions per subset). Direct reuse falls below the floor on all three QA subsets, and position alignment alone brings it back close to the ceiling. The loss therefore comes from positions, and position alignment removes it without recomputation. Room for repair appears when several separately cached blocks are
combined.

\section{Method Matrix}
\label{app:matrix}
Table~\ref{tab:matrix} gives the full method matrix behind the class glyphs of the main boards and the supplementary report experiment. Every method follows its
paper and official implementation, with minimal modifications to run on our models and inputs; trained methods use their published
training recipes. Methods that cannot run in our setting keep a row, with the reason in a footnote.

\begin{table}[h]
\caption{Method matrix. Classes: \cls{1}~selective recomputation,
\cls{2}~attention calibration, \cls{3}~position alignment,
\cls{4}~anchor-delta reuse, \cls{5}~latent communication, \cls{6}~fine-tuned
repair, \cls{7}~trained soft-token adapter, \cls{8}~repair of generated KV, \cls{9}~compression under reuse. Prefix caching is not listed: on our inputs it equals full recomputation (Appendix~\ref{sec:ecosystem}).
Cell legend: \cmark~=~evaluated; \xmark~=~not evaluated (reason in the lettered footnote); \nacell~=~does not apply to this setting; ref.~=~reference, not ranked.}
\label{tab:matrix}
\begin{center}
\scriptsize
\setlength{\tabcolsep}{3pt}
\begin{tabular}{@{}P{4.8cm}llcc@{}}
\toprule
Method & Class & Train & Retr.\ Evid. & \hdr{Reports\\(supp.)} \\
\midrule
CacheBlend \citep{cacheblend2025} & \cls{1} & none & \cmark & \cmark \\
LegoLink (EPIC) \citep{epic2025} & \cls{1} & none & \cmark & \cmark \\
DroidSpeak \citep{droidspeak2025} & \cls{1} & none & \xmark$^{f}$ & \xmark$^{f}$ \\
\midrule
APE \citep{ape2025} & \cls{2} & none & \cmark & \cmark \\
\midrule
Position alignment & \cls{3} & none & ref. & ref. \\
\midrule
KVCOMM \citep{kvcomm2025} & \cls{4} & none & \cmark$^{a}$ & \nacell$^{a}$ \\
\midrule
LatentMAS \citep{latentmas2026} & \cls{5} & none & \nacell$^{d}$ & \nacell$^{d}$ \\
\midrule
Block-Attention \citep{blockattention2024} & \cls{6} & 8.2 GPU-h & \cmark & \cmark \\
TurboRAG \citep{turborag2024} & \cls{6} & training & \nacell$^{b}$ & \nacell$^{b}$ \\
KVLink \citep{kvlink2025} & \cls{6} & training & \xmark$^{c}$ & \xmark$^{c}$ \\
\midrule
KVPacket \citep{kvpacket2026} & \cls{7} & adapter & \cmark & \cmark \\
\midrule
RelayCaching \citep{relaycaching2026} & \cls{8} & none & \nacell$^{e}$ & \cmark \\
\midrule
4-bit cache quantization & \cls{9} & none & \cmark & \cmark \\
SnapKV / Knorm (eviction) \citep{snapkv2024,knorm2024} & \cls{9} & none & \cmark & \cmark \\
TOVA / StreamingLLM (eviction) \citep{tova2024,streamingllm2024} & \cls{9} & none & \cmark & \cmark \\
\bottomrule
\end{tabular}
\end{center}
\vspace{2pt}
{\scriptsize
\par\noindent $^{a}$~KVCOMM stores offsets from earlier requests and reuses a cache when its document appears again. The leaderboard runs it on the questions in order, where $45\%$, $25\%$, and $0\%$ of
scientific-paper, single-document, and multi-hop questions reuse the
cache (Table~\ref{tab:leaderboard}).
The fixed reports do not repeat, so it does not apply to Agent Reports.
\par\noindent $^{b}$~No released training code.
\par\noindent $^{c}$~Not run at the training scale of its published recipe.
\par\noindent $^{d}$~LatentMAS changes what agents exchange (latent states instead of text) rather than repairing a cache.
\par\noindent $^{e}$~RelayCaching repairs the KV an agent produced while generating its output; Retrieved Evidence has no generated text to hand over.
\par\noindent $^{f}$~No released code.}
\end{table}

\section{Latency and Memory by Dataset}
\label{app:frontiers}
These dataset-level views complement the main-text Pareto overview in Figure~\ref{fig:pareto}. They use the Qwen3-8B board; Table~\ref{tab:memory} in Section~\ref{sec:re-board} averages KV memory over all five models.

\begin{figure}[h]
\begin{center}
\includegraphics[width=\linewidth]{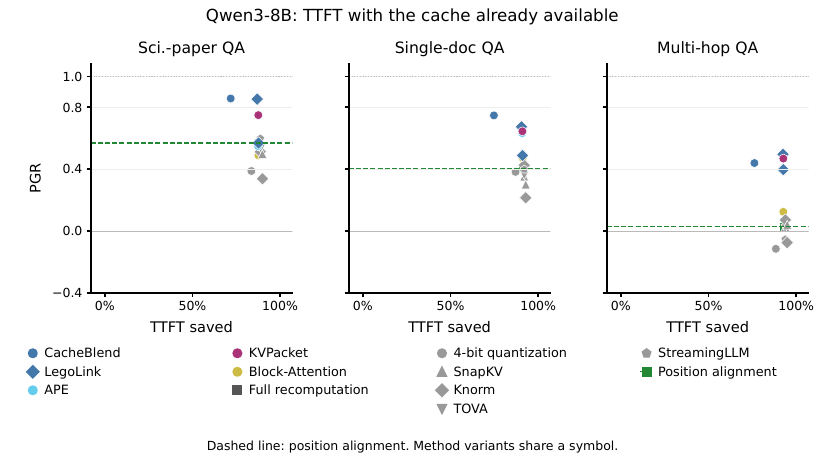}
\end{center}
\caption{On Qwen3-8B, methods that recompute more tokens save less TTFT. Repairs that recompute no document tokens save $.87$--$.93$ of TTFT; CacheBlend, run at a $15\%$ recomputation budget, saves $.72$--$.76$. Per-request
time to first token with the cache already available, all methods on one backend against the same full recomputation. Color = method class as in
Figure~\ref{fig:boards}; configurations of a method share a symbol.}
\label{fig:frontier-runtime}
\end{figure}

\subsection{Break-even with deployment numbers}
\label{app:amortization}
A deployer can compute the break-even point with their own numbers: a cache pays
for itself after
$T_{\mathrm{build}}/(T_{\mathrm{full}} - T_{\mathrm{method}} - B_{\mathrm{method}}/\text{bandwidth})$
reads, where $T_{\mathrm{build}}$ is the time to build the cache and bandwidth is the speed of loading it. The benchmark reports $T_{\mathrm{full}}$, $T_{\mathrm{method}}$, and $B_{\mathrm{method}}$ (\S\ref{sec:frontiers}); the build time and the bandwidth depend on the deployment. How often a workload reads a cache decides how soon the cache reaches this count. Prior work has measured both read patterns and loading speed, in serving engines and on traces. RAGCache reports skewed retrieval: on its MMLU workload, the top $3\%$ of documents serve $60\%$ of requests. Its cache hits are up to $3.9\times$ faster than full prefill, counting the host-to-GPU transfer \citep{ragcache2024}. CacheBlend loads one layer's KV cache for a 4K-token Llama-7B context from an NVMe SSD in $16$\,ms. It overlaps selective recomputation with this loading and reports a $2.2$--$3.3\times$ TTFT reduction against full recomputation in its serving stack \citep{cacheblend2025}. KVCOMM measures $68$--$88\%$ of
tokens reusable across two to five agents in multi-agent RAG, math, and
coding pipelines \citep{kvcomm2025}, and DroidSpeak reports $3.1\times$
faster prefill across two GPU nodes over InfiniBand \citep{droidspeak2025}.
Our per-request savings with the cache already available are in Figure~\ref{fig:frontier-runtime}. A cache streamed from slower storage adds $B_{\mathrm{method}}/\text{bandwidth}$ to every read, and Table~\ref{tab:memory} gives $B_{\mathrm{method}}$.

\begin{figure}[h]
\begin{center}
\includegraphics[width=\linewidth]{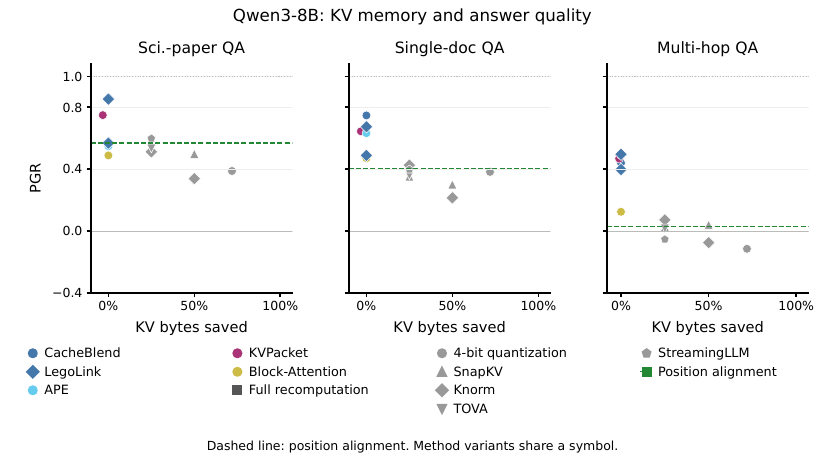}
\end{center}
\caption{Memory cost: only compression (gray) saves KV memory, at a cost in quality; every other method stores the full cache, and KVPacket stores about $3\%$ extra. Bytes are
measured on the actual tensors, so 4-bit quantization saves a measured
$71.9\%$ rather than the nominal $75\%$. Color = method class;
configurations of a method share a symbol. The dashed line marks position alignment.}
\label{fig:frontier-memory}
\end{figure}
\FloatBarrier

\FloatBarrier
\section{Dataset Cards}
\label{app:data}

\textbf{Retrieved Evidence subsets.} The question sets contain 1,000
Qasper questions, 150 MultiFieldQA questions, and 1,000 HotpotQA
questions: 2,150 unique questions in total, covering scientific-paper,
single-document, and multi-hop QA. LongBench~v1 provides 200, 150, and
200 of them. The added Qasper questions come from the official Qasper
development and test splits, formatted with LongBench's own procedure,
which reproduces LongBench's Qasper samples exactly; the 1,000 questions
cover 547 papers. HotpotQA keeps LongBench's 200 questions unchanged. The added questions come from the HotpotQA development split
(distractor setting) and are assembled as in LongBench: each passage is
the full Wikipedia article of a supporting or distractor title, taken
from a fixed Wikipedia dump, under a length budget matched to
LongBench's. We drew 800 of them with a fixed seed from the candidates
whose supporting titles are in the dump and that fit the budget.
MultiFieldQA uses all 150 LongBench questions. No question appears in
the training data of the trained methods: the Block-Attention
fine-tuning pool, HotpotQA train, and MuSiQue train; KVPacket's other training data are synthetic. We use natural passage boundaries
where provided and $512$-token windows otherwise, with an input cap of
29K tokens. Context value (ceiling$-$floor) on the 8B reference board is
$.326/.361/.338$ F1.

\begin{figure}[t]
\centering
\includegraphics[width=\linewidth]{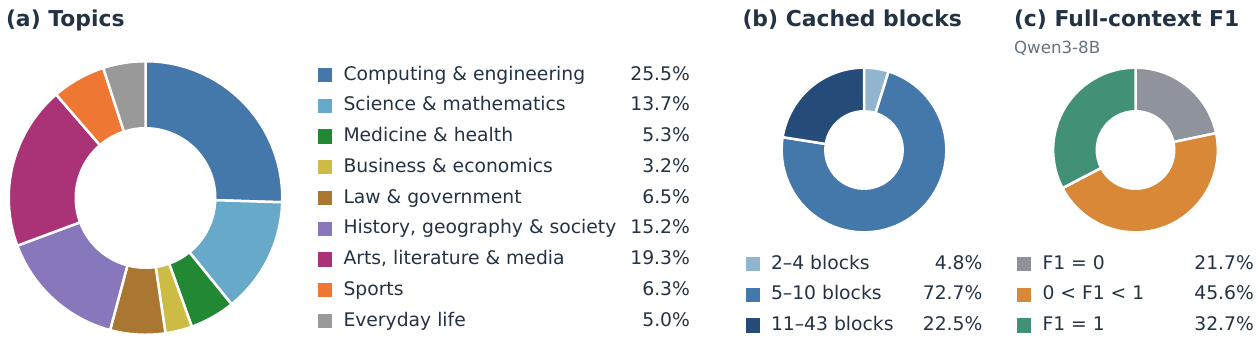}
\caption{\textbf{Dataset composition.}
\textbf{(a)}~Topic distribution.
\textbf{(b)}~Number of independently cached blocks per question.
\textbf{(c)}~Answer F1 after full recomputation with Qwen3-8B.
}
\label{fig:dataset-coverage}
\end{figure}

\textbf{Dataset composition.} Figure~\ref{fig:dataset-coverage} shows the topics, the number of blocks per question, and the full-recomputation F1 of the question sets. Topics fall into nine categories. Each question has 2--43 blocks, each a passage or a 512-token window encoded on its own. The F1 distribution uses
Qwen3-8B's full-recomputation answers.

\textbf{Supplementary reports.} The report construction, its floor and ceiling, and the results appear together in
Appendix~\ref{app:agent-reports}.

\subsection{Context value and room for repair}
\label{sec:admission}
Two measured quantities describe each board and subset (Table~\ref{tab:headroom}, \S\ref{sec:confirmation}). The context is \emph{worth} ceiling$-$floor. Position alignment \emph{leaves} the share $(\mathrm{ceiling}-S_{\mathrm{align}})/(\mathrm{ceiling}-\mathrm{floor})$ to repair, where $S_{\mathrm{align}}$ is the score of position alignment. If a model answers nearly as well without the documents, the context is worth little. If a model reads separately encoded caches well, little is left to repair. In both cases every method scores close to position alignment, so the methods cannot be told apart. The table also helps read the results: repair matters where both numbers are large and fades where little is left.
The floor, ceiling, position alignment, and prefix caching are reference points and are not ranked (Appendix~\ref{sec:ecosystem}).

\section{Run Settings}
\label{app:settings}
Table~\ref{tab:settings} lists the settings shared by every method.

\begin{table}[h]
\caption{\textbf{Every method runs on the same models, inputs, and
scoring, and is timed on one shared backend.} Cross-checkpoint producers
are listed in Table~\ref{tab:pairs}.}
\label{tab:settings}
\begin{center}
\small
\begin{tabular}{@{}lP{0.76\linewidth}@{}}
\toprule
Setting & Value \\
\midrule
Models & Qwen3-4B, Qwen3-8B, Qwen3-14B, Qwen3-30B-A3B-Instruct-2507
(mixture of experts), and Llama-3.1-8B-Instruct; on each board the model
writes and reads its own caches \\
Input & up to 29K tokens; natural passage boundaries where provided,
otherwise 512-token windows; 2--43 cached blocks per question \\
Scoring & LongBench's official token-level F1; PGR against each board's
floor and ceiling \\
Timing & time to first token with the cache already available, on NVIDIA H100 GPUs
with one shared vLLM backend; each method and full recomputation run on
the same model and GPU \\
\bottomrule
\end{tabular}
\end{center}
\end{table}

\section{Notes on the Per-Board Results}
\label{app:confirmation}
The per-board tables (Tables~\ref{tab:board-8B}--\ref{tab:board-30B}) show which gains hold on every model and which depend on the model. CacheBlend significantly improves scientific-paper
and single-document QA on all five models. KVPacket and APE
significantly improve multi-hop QA on all five models, and Block-Attention
on four. LegoLink improves
single-document and multi-hop QA on the four Qwen boards. KVCOMM gains on single-document and multi-hop QA mainly by
recomputing the questions whose documents it has not seen. The
supplementary report
results are discussed in Appendix~\ref{app:agent-reports}.

Fine-tuning with Block-Attention's published recipe lowers the fine-tuned model's own ceiling on four of the five models (scientific-paper / single-document / multi-hop QA: 4B $-.20/-.12/-.33$, Llama $-.20/-.14/-.39$, 30B $-.10/-.05/-.17$; 14B $-.08$ on scientific-paper QA only). The 8B ceiling does not drop. With the fine-tuned model's own floor and ceiling, the multi-hop gap (ceiling$-$floor) falls below $.10$ on 4B and Llama. We therefore score every Block-Attention result against the base model's floor and ceiling.

\section{Robustness: Does a Repair Stay Effective When the Setting Changes?}
\label{sec:robustness}
Methods respond differently to changes of model and producer.
A mean over boards hides whether a repair works
everywhere or only somewhere, so Table~\ref{tab:robustness} reports the
worst case over settings beside the means, as WILDS reports worst-group
accuracy beside average accuracy \citep{wilds2021}. We ask whether a repair's effect survives two changes of setting. \emph{Across models}: a method's gain on a board is its PGR
minus that of position alignment on the same board; the table gives the worst
gain over the five boards and the number of boards on which the gain
is positive. \emph{Across producers}: in each of $18$ tests (the six pairs of Table~\ref{tab:pairs} on three subsets), we take the paired change in the method's own F1 when another checkpoint of the same family writes the cache. The table gives the mean and worst change with
95\% intervals and the number of tests whose interval lies below zero.
CacheBlend improves
single-document QA on every model (worst gain $+.20$ PGR) and multi-hop
QA on four of five. KVPacket improves on position alignment on every model and QA subset. It is also hurt most often by a producer change ($12$ of $18$ tests significantly below zero; worst change $-.146$ F1). APE
improves multi-hop QA on all five models, with a worst gain of $+.11$
PGR. LegoLink with $k{=}0$ also has one large producer drop ($-.151$ F1).
A cache written by a model from a different pretraining run lowers every method (Table~\ref{tab:diff-pretrain}).

\begin{table}[!ht]
\caption{Retrieved Evidence: worst-case results across the tested settings.
Left: the worst
gain over position alignment across the five boards (PGR units);
superscript = boards with a positive gain / all boards; \textbf{bold}
= positive on every board and significant on a majority. Right: the mean
and worst paired change of the method's own F1 when a
same-family checkpoint writes the cache ($\pm$ 95\% interval), and the
number of the $18$ tests whose interval lies below zero. KVCOMM is left out: it recomputes most questions, so these columns would not measure reuse.}
\label{tab:robustness}
\begin{center}
\scriptsize
\setlength{\tabcolsep}{2.5pt}
\begin{tabular}{@{}lrrrrr@{}}
\toprule
& \multicolumn{2}{c}{Worst gain over position alignment} & \multicolumn{3}{c}{Producer change (F1)} \\
\cmidrule(lr){2-3}\cmidrule(lr){4-6}
Method & Single-doc & Multi-hop & mean & worst & tests sig. below \\
\midrule
\cls{3}~Position alignment & -- & -- & $-$.008$\pm$.005 & $-$.035$\pm$.032 & 4/18 \\
\midrule
\cls{1}~CacheBlend (15\% recompute) & \textbf{.202$^{5/5}$} & $-$.077$^{4/5}$ & $-$.005$\pm$.005 & $-$.025$\pm$.030 & 0/18 \\
\cls{1}~LegoLink ($k{=}2$, $\approx$0.4\% recompute) & $-$.009$^{4/5}$ & $-$.016$^{4/5}$ & $-$.011$\pm$.005 & $-$.037$\pm$.031 & 5/18 \\
\cls{1}~LegoLink ($k{=}0$, 0\% recompute) & $-$.014$^{4/5}$ & $-$.028$^{4/5}$ & $-$.019$\pm$.006 & $-$.151$\pm$.024 & 8/18 \\
\cls{2}~APE & .027$^{5/5}$ & \textbf{.110$^{5/5}$} & $-$.006$\pm$.005 & $-$.040$\pm$.037 & 6/18 \\
\cls{6}~Block-Attention & $-$.088$^{2/5}$ & \textbf{.057$^{5/5}$} & $-$.004$\pm$.004 & $-$.034$\pm$.022 & 6/18 \\
\cls{7}~KVPacket (0\% recompute) & \textbf{.109$^{5/5}$} & \textbf{.201$^{5/5}$} & $-$.039$\pm$.007 & $-$.146$\pm$.023 & 12/18 \\
\cls{9}~Quant.\ 4-bit & $-$.140$^{0/5}$ & $-$.706$^{0/5}$ & $-$.011$\pm$.005 & $-$.025$\pm$.033 & 2/18 \\
\cls{9}~SnapKV ($r{=}.25$) & $-$.105$^{0/5}$ & $-$.014$^{1/5}$ & $-$.007$\pm$.005 & $-$.038$\pm$.028 & 2/18 \\
\cls{9}~SnapKV ($r{=}.5$) & $-$.141$^{0/5}$ & $-$.040$^{3/5}$ & $-$.005$\pm$.004 & $-$.038$\pm$.017 & 4/18 \\
\cls{9}~Knorm ($r{=}.25$) & $-$.168$^{2/5}$ & .041$^{5/5}$ & $-$.016$\pm$.005 & $-$.034$\pm$.037 & 8/18 \\
\cls{9}~Knorm ($r{=}.5$) & $-$.342$^{0/5}$ & $-$.142$^{2/5}$ & .000$\pm$.004 & $-$.007$\pm$.016 & 0/18 \\
\cls{9}~TOVA ($r{=}.25$) & $-$.093$^{0/5}$ & $-$.042$^{2/5}$ & $-$.006$\pm$.005 & $-$.037$\pm$.036 & 4/18 \\
\cls{9}~StreamingLLM ($r{=}.25$) & $-$.113$^{0/5}$ & $-$.307$^{0/5}$ & .001$\pm$.004 & $-$.030$\pm$.036 & 0/18 \\
\bottomrule
\end{tabular}

\end{center}
\end{table}

\begin{table}[!ht]
\caption{A cache written by a model from a different pretraining run
lowers the F1 of every method. The receiver is
Qwen3-30B-A3B-Instruct-2507 and the cache is written by
Qwen3-Coder-30B-A3B-Instruct, which has the same architecture. Entries
are the paired change in F1 relative to the receiver's own
cache, on 100 questions per subset; $^{\downarrow}$ = the 95\% interval
lies below zero.}
\label{tab:diff-pretrain}
\begin{center}
\scriptsize
\begin{tabular}{@{}lrrr@{}}
\toprule
Method & Sci.-paper QA & Single-doc QA & Multi-hop QA \\
\midrule
\cls{3}~Position alignment & $-$.165$^{\downarrow}$ & $-$.183$^{\downarrow}$ & $-$.214$^{\downarrow}$ \\
\midrule
\cls{1}~LegoLink ($k{=}2$, $\approx$0.4\% recompute) & $-$.254$^{\downarrow}$ & $-$.224$^{\downarrow}$ & $-$.309$^{\downarrow}$ \\
\cls{1}~LegoLink ($k{=}0$, 0\% recompute) & $-$.275$^{\downarrow}$ & $-$.230$^{\downarrow}$ & $-$.320$^{\downarrow}$ \\
\cls{2}~APE & $-$.255$^{\downarrow}$ & $-$.195$^{\downarrow}$ & $-$.190$^{\downarrow}$ \\
\cls{6}~Block-Attention & $-$.129$^{\downarrow}$ & $-$.179$^{\downarrow}$ & $-$.005 \\
\cls{7}~KVPacket (0\% recompute) & $-$.141$^{\downarrow}$ & $-$.217$^{\downarrow}$ & $-$.287$^{\downarrow}$ \\
\cls{9}~Quant.\ 4-bit & $-$.197$^{\downarrow}$ & $-$.183$^{\downarrow}$ & $-$.138$^{\downarrow}$ \\
\cls{9}~SnapKV ($r{=}.25$) & $-$.160$^{\downarrow}$ & $-$.199$^{\downarrow}$ & $-$.210$^{\downarrow}$ \\
\cls{9}~SnapKV ($r{=}.5$) & $-$.164$^{\downarrow}$ & $-$.158$^{\downarrow}$ & $-$.235$^{\downarrow}$ \\
\cls{9}~Knorm ($r{=}.25$) & $-$.164$^{\downarrow}$ & $-$.144$^{\downarrow}$ & $-$.142$^{\downarrow}$ \\
\cls{9}~Knorm ($r{=}.5$) & $-$.171$^{\downarrow}$ & $-$.051$^{\downarrow}$ & $-$.116$^{\downarrow}$ \\
\cls{9}~TOVA ($r{=}.25$) & $-$.153$^{\downarrow}$ & $-$.156$^{\downarrow}$ & $-$.186$^{\downarrow}$ \\
\cls{9}~StreamingLLM ($r{=}.25$) & $-$.128$^{\downarrow}$ & $-$.143$^{\downarrow}$ & $-$.128$^{\downarrow}$ \\
\bottomrule
\end{tabular}
\end{center}
\end{table}

\section{Relation to Prefix Caching}
\label{sec:ecosystem}

On our boards, prefix caching coincides with full recomputation, the reference point of the quality--cost plots. Prefix caching is lossless because it reuses KV only after the same text. On our inputs, a prefix cache matches only the fixed instruction at the start of each prompt, so a single request saves no document computation. We report it in two ways. Per request, it keeps full quality and saves no document computation. Over the full sequence of questions, its hit rate is $0.6\%$ to $21.4\%$ per subset. Hits beyond the instruction come from prefixes that samples share, and its quality does not differ significantly from full recomputation. Prefix caching and repair methods address different cases: prefix caching saves work when the preceding text is unchanged, and repair methods recover quality when it changes.

Serving engines and repair methods are complementary. CacheBlend already ships as a cache-layer plugin for these serving engines, and other repair classes could ship the same way. We read each method against full recomputation: how much cost it saves and how much quality it keeps.

\section{Open Platform and Submission Workflow}
\label{app:platform}
\label{sec:framework}
New methods use the same inputs and scoring rules as existing entries. The public repository provides the evaluation package, fixed inputs,
reference scores, and leaderboard data (repository
\repourl; leaderboard \leaderboardurl; data \dataseturl). The package installs from the repository with pip (\texttt{pip install -e .}).
\paragraph{Adding a method.}
A method implements three calls: \texttt{setup} loads the model or
engine once; \texttt{build\_cache} encodes one block without seeing the question; \texttt{answer} reuses the caches and returns an answer with
cost measurements. The cost fields record four numbers: document tokens recomputed, document tokens full recomputation would compute (both counted per layer), KV bytes, and time to first token.
An example driver implements position alignment.
Every method uses the same fixed inputs and reference scores.

\paragraph{Scoring and submission.}
\texttt{kva validate} checks the result format and fixed sample IDs.
\texttt{kva score} re-scores the predictions on CPU and computes PGR
from the released references. \texttt{kva submit} prepares a result file and a
method card for a pull request. The card documents the method and how its costs were measured.
Maintainers check the model and prompt settings, reference answers,
recomputed scores, and documented costs before updating the leaderboard data.

\paragraph{Leaderboard and release coverage.}
The leaderboard page is built from a versioned result file. Speed is the share of TTFT saved, as in the main results, and users can set the weight between quality and speed.
The public leaderboard shows the results of this paper for the
five receiver models. The package includes the floor and ceiling of every
model and subset, so a full run can be scored as PGR. The package generates answers for Retrieved Evidence and can also score Agent Reports answers.
\section{Cross-Checkpoint Supplement}
\label{app:crosscheckpoint}
Table~\ref{tab:llama-cross} gives each method's F1 change on Llama-3.1-8B-Instruct when the base model writes the cache. Position alignment loses $.026$ on scientific-paper QA and $.035$ on single-document QA, both significant, and APE gains $.051$ on multi-hop QA.

\begin{table}[h]
\caption{\textbf{Every cross-checkpoint pair keeps the architecture and
tokenizer and changes only the weights.} The receiver is the board's
model; the producer writes the caches.}
\label{tab:pairs}
\begin{center}
\footnotesize
\begin{tabular}{@{}lll@{}}
\toprule
Receiver (board) & Producer & Relation \\
\midrule
Qwen3-8B & Qwen3-8B-Base & base predecessor \\
Qwen3-8B & DeepSeek-R1-0528-Qwen3-8B & distilled sibling \\
Qwen3-4B & Qwen3-4B-Base & base predecessor \\
Qwen3-14B & Qwen3-14B-Base & base predecessor \\
Llama-3.1-8B-Instruct & Llama-3.1-8B & base predecessor \\
Qwen3-30B-A3B-Instruct-2507 & Qwen3-30B-A3B-Thinking-2507 & post-trained sibling \\
\bottomrule
\end{tabular}
\end{center}
\end{table}

\begin{table}[h]
\caption{Without repair, changing the context or the producer lowers F1. Scores are F1 on Retrieved Evidence with Qwen3-8B as the receiver; where the producer differs, the two values are for the base predecessor and the distilled sibling (Table~\ref{tab:pairs}).}
\label{tab:factorial-results}
\begin{center}
\small
\begin{tabular}{@{}llcc@{}}
\toprule
Context & Producer weights & HotpotQA & Qasper \\
\midrule
Same & Same as receiver & $.559$ & $.417$ \\
Same & Different & $.484/.527$ & $.353/.410$ \\
Different & Same as receiver & $.394$ & $.397$ \\
Different & Different & $.361/.315$ & $.381/.344$ \\
\bottomrule
\end{tabular}
\end{center}
\end{table}

\begin{table}[h]
\caption{Llama-3.1-8B base $\to$ Instruct: paired F1 change after replacing
the cache producer. Each method is compared with its own same-model
result on the same questions. $*$ marks a paired 95\% confidence
interval excluding zero.}
\label{tab:llama-cross}
\begin{center}
\small
\setlength{\tabcolsep}{4pt}
\begin{tabular}{@{}lrrr@{}}
\toprule
Method & Sci.-paper QA & Single-doc QA & Multi-hop QA \\
\midrule
\cls{3}~Position alignment & $-$.026* & $-$.035* & $-$.009 \\
\midrule
\cls{1}~CacheBlend (15\% recompute) & $-$.010 & $-$.025 & $-$.005 \\
\cls{1}~LegoLink ($k{=}2$, $\approx$0.4\% recompute) & $-$.023* & $-$.037* & $-$.003 \\
\cls{1}~LegoLink ($k{=}0$, 0\% recompute) & .003 & $-$.039* & $-$.023* \\
\cls{2}~APE & .015 & $-$.040* & .051* \\
\cls{6}~Block-Attention & $-$.011* & $-$.034* & $-$.013* \\
\cls{7}~KVPacket (0\% recompute) & $-$.013 & $-$.068* & $-$.027* \\
\cls{9}~Quant.\ 4-bit & $-$.023* & $-$.025 & .000 \\
\cls{9}~SnapKV ($r{=}.25$) & $-$.019* & $-$.038* & .009 \\
\cls{9}~SnapKV ($r{=}.5$) & $-$.011 & $-$.037* & .017 \\
\cls{9}~Knorm ($r{=}.25$) & $-$.021* & $-$.034 & $-$.001 \\
\cls{9}~Knorm ($r{=}.5$) & $-$.006 & $-$.006 & $-$.006 \\
\cls{9}~TOVA ($r{=}.25$) & $-$.018* & $-$.037* & $-$.003 \\
\cls{9}~StreamingLLM ($r{=}.25$) & $-$.008 & $-$.030 & .038* \\
\bottomrule
\end{tabular}

\end{center}
\end{table}

\subsection{How producer caches are injected}
For a producer change, each method receives KV written by another checkpoint of the same architecture instead of the receiver's own. With the receiver's own KV, the same code reproduces every method's original outputs, so the swap itself changes nothing.

\subsection{Answers of the trained adapter under a producer change}
Where KVPacket's adapter loses quality under a different producer, its answers remain fluent but get facts wrong (for example, a wrong
actress, a flipped yes/no, or a year attached to the wrong event).

\subsection{Retraining the adapter per producer}
\label{app:adapter-retrain}
To test whether retraining alone restores quality, we retrained the KVPacket adapter with its published recipe and budget on caches written by each new producer, with the receiver fixed. Three adapters were evaluated on the same six tests under the same setup: the original adapter of the main board, the
adapter retrained on the producer under test, and the adapter
retrained on the other new producer (cross). The retrained adapter shows no significant difference from the same-model baseline in any of the six tests. It is significantly above the original adapter in the five tests where the original adapter had lost quality, by up to $.25$ F1. The cross-trained adapter recovers part of the loss with the distilled sibling as producer and none with the base predecessor, and stays below the baseline in every test. Each retraining took about $20$ GPU-hours.

\begin{table}[h]
\caption{KVPacket adapter under a different producer (receiver Qwen3-8B; F1). Same-model baseline: $.407/.401/.447$
on Sci.-paper / Single-doc / Multi-hop QA. Bold = paired 95\% CI against
the same-model baseline excludes zero.}
\label{tab:adapter-retrain}
\begin{center}
\small
\begin{tabular}{@{}llccc@{}}
\toprule
Producer & Adapter & Sci.-paper QA & Single-doc QA & Multi-hop QA \\
\midrule
\multirow{3}{*}{Distilled sibling} & original & $\mathbf{.340}$ & $\mathbf{.292}$ & $\mathbf{.248}$ \\
 & retrained on this producer & .398 & .374 & .497 \\
 & cross-trained (on the base producer) & .381 & .353 & .398 \\
\midrule
\multirow{3}{*}{Base predecessor} & original & .369 & $\mathbf{.347}$ & .467 \\
 & retrained on this producer & .411 & .427 & .442 \\
 & cross-trained (on the sibling) & .371 & .346 & .408 \\
\bottomrule
\end{tabular}
\end{center}
\end{table}

\section{Leaderboard Score Under Other Weights}
\label{app:score}
Table~\ref{tab:score-presets} lists the leaderboard score of
\S\ref{sec:leaderboard} under three quality weights: $0.7$ (quality first),
$0.5$ (the default), and $0.3$ (speed first). KVPacket and LegoLink lead under all three weights, and repair methods take the top five places. CacheBlend, which gives up speed for quality, ranks third at $0.7$ and eleventh at $0.3$; KVCOMM ranks sixth at $0.7$ and last at $0.5$ and $0.3$.

\begin{table}[h]
\caption{Leaderboard score under three quality weights; $0.5$ is the
default of Table~\ref{tab:leaderboard}. Rows are ordered by the default; ranks under each weight are in parentheses. $^{*}$Recomputes the questions whose documents it has not seen.}
\label{tab:score-presets}
\begin{center}
\scriptsize
\setlength{\tabcolsep}{3pt}
\begin{tabular}{@{}lrrr@{}}
\toprule
Method & quality first ($0.7$) & balanced ($0.5$) & speed first ($0.3$) \\
\midrule
\cls{7}~KVPacket (0\% recompute) & .746 (1) & .789 (1) & .832 (1) \\
\cls{1}~LegoLink ($k{=}2$, $\approx$0.4\% recompute) & .733 (2) & .778 (2) & .824 (2) \\
\cls{2}~APE & .689 (4) & .748 (3) & .807 (3) \\
\cls{1}~LegoLink ($k{=}0$, 0\% recompute) & .672 (5) & .736 (4) & .801 (4) \\
\cls{1}~CacheBlend (15\% recompute) & .713 (3) & .720 (5) & .726 (11) \\
\cls{6}~Block-Attention & .598 (7) & .684 (6) & .769 (5) \\
\cls{9}~SnapKV ($r{=}.25$) & .577 (8) & .671 (7) & .766 (6) \\
\cls{9}~Knorm ($r{=}.25$) & .576 (9) & .671 (8) & .765 (7) \\
\cls{9}~TOVA ($r{=}.25$) & .567 (10) & .664 (9) & .762 (9) \\
\cls{9}~SnapKV ($r{=}.5$) & .555 (11) & .659 (10) & .762 (8) \\
\cls{9}~StreamingLLM ($r{=}.25$) & .544 (12) & .648 (11) & .752 (10) \\
\cls{9}~Quant.\ 4-bit & .490 (13) & .596 (12) & .702 (13) \\
\cls{9}~Knorm ($r{=}.5$) & .466 (14) & .595 (13) & .724 (12) \\
\cls{4}~KVCOMM$^{*}$ & .614 (6) & .490 (14) & .365 (14) \\
\bottomrule
\end{tabular}

\end{center}
\end{table}

\end{document}